\documentclass[11pt]{article}

\usepackage{acl}

\usepackage{times}
\usepackage{latexsym}

\usepackage[T1]{fontenc}
\usepackage[utf8]{inputenc}

\usepackage{microtype}

\usepackage{inconsolata}

\usepackage{graphicx}
\usepackage{booktabs}
\usepackage{amsmath}

\newcommand{\rot}[1]{%
    \rotatebox[origin=c]{35}{\strut #1}%
}

\title{DeferKV: Rethinking Eviction Timing for One-Shot KV Cache Compression}

\author{
Zhe Wang$^{1,2}$ \quad
Jiakai Li$^{1,2}$ \quad
Yujia Sun$^{1,2}$ \quad
Rongzheng Wang$^{1,2}$ \quad
Shuang Liang$^{1,2}$\thanks{Corresponding author: \texttt{shuangliang@uestc.edu.cn}} \\
$^{1}$University of Electronic Science and Technology of China, Chengdu, China \\
$^{2}$Ubiquitous Intelligence and Trusted Services Key Laboratory of Sichuan Province, Chengdu, China
}

\begin{document}
\maketitle
\begin{abstract}
Long-context large language models (LLMs) have demonstrated strong capabilities across a wide range of tasks, but the growing KV cache
introduces substantial memory and inference overhead. Existing one-shot KV cache compression methods typically commit to irreversible eviction immediately after prefill, before any signal from actual generation becomes available. Our quantitative analysis shows that early queries from the actual generation stage provide attention signals that are more consistent with subsequent decode attention, with the largest single-step gain occurring at the prefill--decode boundary. Based on this observation, we propose DeferKV, which moves the eviction decision from the end of prefill to the first real decoding step and temporally combines prompt-side and decode-side observations, thereby better aligning KV importance estimation with subsequent generation requirements. DeferKV requires no additional
training, draft model, or future-query prediction module, making it simple and easy to deploy.
Experiments on LongBench, RULER, and Needle-in-a-Haystack demonstrate
that DeferKV consistently improves model performance under KV cache
compression while maintaining low inference latency.
\end{abstract}

\section{Introduction}

Large language models (LLMs) with long-context capabilities~\cite{DBLP:journals/corr/abs-2303-08774, DBLP:journals/corr/abs-2505-09388} have demonstrated strong performance in document question answering~\cite{DBLP:conf/naacl/DasigiLBCSG21}, multi-document
summarization~\cite{DBLP:conf/acl/FabbriLSLR19}, code understanding~\cite{DBLP:conf/acl/BaiLZL0HDLZHDTL24}, retrieval-augmented generation~\cite{DBLP:conf/nips/LewisPPPKGKLYR020}, and
complex reasoning~\cite{DBLP:conf/iclr/YaoZYDSN023}. However, long-context inference also introduces
substantial system overhead. In particular, the KV cache has become a
major bottleneck affecting memory consumption and inference throughput~\cite{DBLP:journals/tmlr/0002LTTXCHD0025,
DBLP:journals/corr/abs-2407-18003,DBLP:conf/emnlp/YuanLZCLWLJCX0H24}.
For autoregressive Transformer decoders~
\cite{DBLP:conf/nips/VaswaniSPUJGKP17}, the key and value
representations of historical tokens must be cached at every layer to
avoid redundant computation during generation. As the context length
increases, the memory consumption of the KV cache grows linearly with
the sequence length, making long-context inference more challenging on
memory-constrained devices and in high-concurrency serving scenarios.
Therefore, compressing the KV cache while preserving generation quality
has become an important research problem in efficient LLM inference.

Existing KV cache compression methods often perform one-time KV eviction
immediately after the prefill stage. Representative methods such as
SnapKV~\cite{DBLP:conf/nips/LiHYVLYCLC24} use a small set of queries near the end of the prompt as an
observation window and estimate the importance of historical tokens
based on the attention scores between these queries and historical keys,
thereby selecting the KV tokens to retain. These methods are simple to
implement, require no additional training, and offer favorable
efficiency in practice.

\begin{figure}[t]
    \centering
    \includegraphics[
        width=0.95\columnwidth
    ]{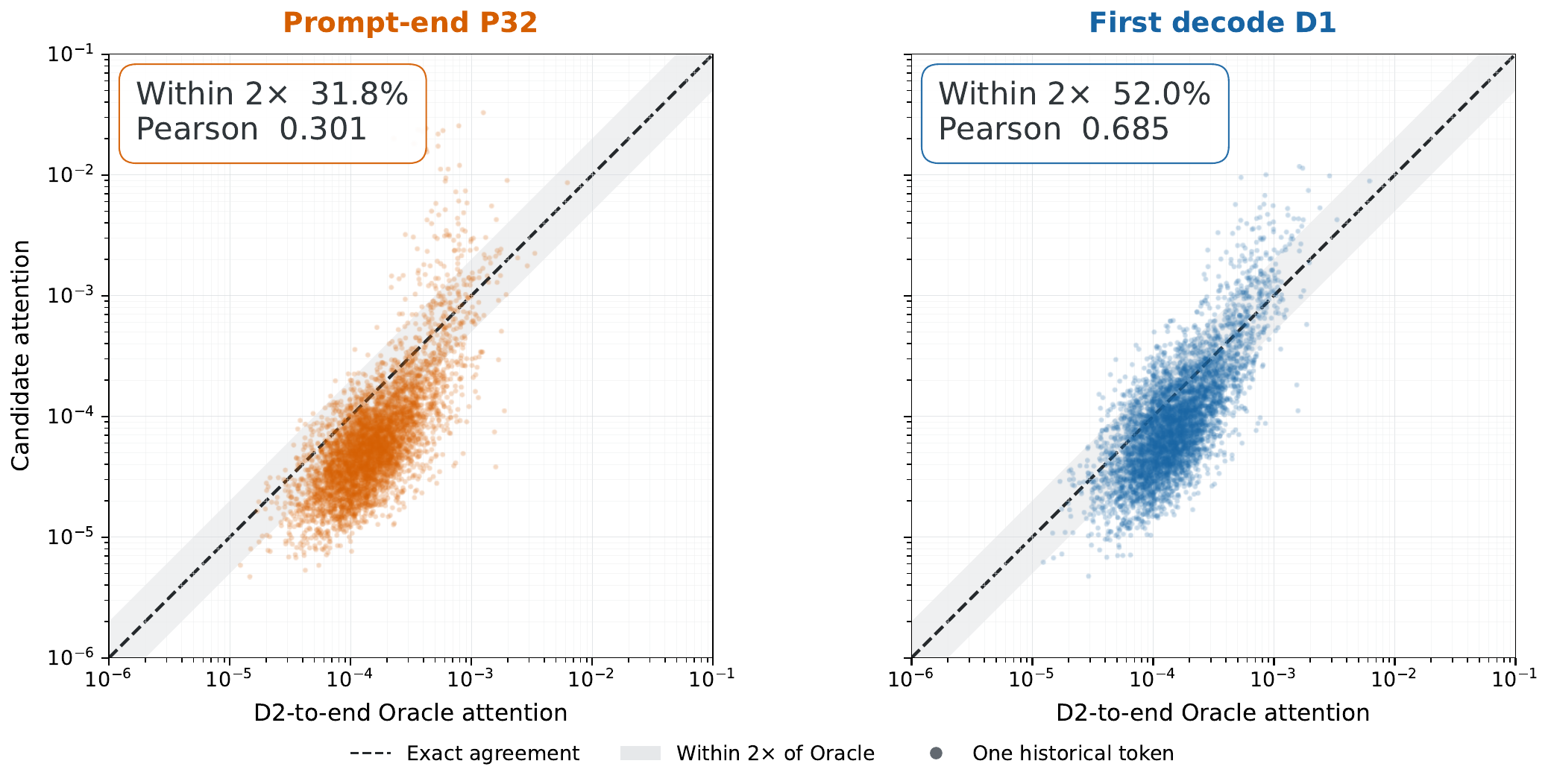}
    \caption{
Consistency of prompt-end and first-decode attention with subsequent
decode attention on GovReport using Qwen3-8B. Each point denotes a
historical KV token. The horizontal axis shows its average attention
from $D_2$ to the end of generation, while the vertical axis shows the
average over the last 32 prefill queries and the
attention from the first decode query. The dashed
diagonal marks exact agreement, and the shaded region indicates values
within a factor of two.
}
    \label{fig:attention_shift}
\end{figure}

However, these methods~\cite{DBLP:conf/nips/LiHYVLYCLC24,DBLP:journals/corr/abs-2502-00299} typically rely on a key assumption: the queries
within the observation window at the end of the prompt are sufficient
to reflect the actual attention requirements of subsequent decoding.
This assumption does not always hold in long-text generation tasks. As shown in Figure~\ref{fig:attention_shift}, more historical KV tokens lie closer to the diagonal when using the first real decode query. Its attention scores are therefore more consistent with subsequent decode attention than those aggregated from the final 32 prefill queries. This suggests that prompt-end attention alone may not fully capture the access patterns observed during subsequent generation. Consequently, estimating KV importance solely from prompt-side queries may lead to suboptimal token selection for later decoding.

Recent methods have begun to address this issue by incorporating information related to future generation into KV token importance estimation. For example, SpecKV~\cite{galim2026draftbased}, LAQ~\cite{DBLP:conf/emnlp/WangJLXXZC25}, and LookaheadKV~\cite{DBLP:journals/corr/abs-2603-10899} construct or
predict future queries or attention patterns to make KV compression more
closely aligned with the requirements of the actual generation stage.
Although these methods alleviate the limitations of prompt-only
observation from different perspectives, they usually require
additional models, auxiliary modules, or extra training procedures and
inevitably face uncertainty introduced by prediction errors.

Based on these observations, we propose a more direct and lightweight
approach. Unlike existing methods that rely solely on prompt-side information, our strategy postpones irreversible KV eviction until the first real decode-side signal becomes available and incorporates a small amount of attention feedback from actual generation. This explicitly aligns the importance estimation of historical tokens with
the requirements of subsequent decoding. We further consider how to
effectively integrate observations from the input and generation stages.
To this end, we introduce a simple yet effective temporal weighting
strategy to improve the reliability and accuracy of importance
estimation.

Our main contributions are summarized as follows:

\begin{itemize}

\item We revisit the timing assumption of existing KV cache compression methods that perform eviction immediately after prefill. We propose a compression approach that postpones irreversible KV eviction until the earliest real decode-side feedback becomes available, thereby better aligning KV selection with subsequent decoding.

\item Our method is simple, direct, and lightweight. It does not require additional training, a draft model, an auxiliary network, or a future-query prediction module. It only requires minor modifications to the existing inference process, enabling straightforward implementation and deployment.

\item Extensive experiments show that our method consistently outperforms
state-of-the-art baselines under constrained KV cache budgets, achieving
relative performance gains of up to 3.58\% on LongBench and 8.03\% on
RULER.

\end{itemize}

\section{Related Work}
\subsection{KV Cache Compression and Token Eviction}

To reduce the memory consumption of the KV cache, existing studies~\cite{DBLP:conf/nips/Zhang00CZC0TRBW23,DBLP:conf/nips/LiuDLWXXKS23} have explored compression by reducing the number of cached tokens. These methods typically retain important Key-Value pairs according to the potential contribution of historical tokens to subsequent generation, while evicting KV entries with lower importance, thereby controlling the cache size during decoding. Prompt-only compression methods usually perform a one-time selection of prompt KV tokens after the prefill stage and before decoding begins~\cite{DBLP:journals/tmlr/0002LTTXCHD0025}. They are generally simple to implement and introduce limited scheduling overhead.

SnapKV~\cite{DBLP:conf/nips/LiHYVLYCLC24} uses an observation window near the end of the prompt to estimate the importance of historical KV tokens and selects clustered important KV positions within each attention head. PyramidKV~\cite{DBLP:journals/corr/abs-2406-02069} further considers layer-wise budget allocation and assigns larger cache budgets to lower layers and smaller budgets to higher layers according to their different information aggregation characteristics. Ada-KV~\cite{DBLP:conf/nips/FengLCXZ25} focuses on the importance differences across attention heads and adaptively allocates more cache capacity to heads that have a greater influence on the model output. ChunkKV~\cite{DBLP:journals/corr/abs-2502-00299} treats semantic chunks
rather than individual tokens as the basic compression units and
introduces layer-wise index reuse to preserve contextual integrity while
reducing selection overhead.

These methods rely primarily on prompt-side observations and may not
fully reflect the actual attention requirements of subsequent decoding.

\subsection{Future-Aware KV Cache Compression}

To alleviate the limitations of prompt-only observation, recent studies have begun to introduce future-aware signals. These methods improve KV selection by predicting or approximating future queries, making the compression strategy more closely aligned with the requirements of the actual generation stage.

SpecKV~\cite{galim2026draftbased} uses a lightweight draft model to generate approximate future tokens and employs these lookahead tokens to help the target model estimate the importance of prompt KV pairs for subsequent generation.

LAQ~\cite{DBLP:conf/emnlp/WangJLXXZC25} performs low-budget lookahead decoding after prefill to generate a small number of pseudo-response tokens. It then caches their corresponding query states as a Q-Cache and uses them to re-evict the KV cache.

LookaheadKV~\cite{DBLP:journals/corr/abs-2603-10899} further introduces learnable lookahead tokens and combines them with LoRA~\cite{DBLP:conf/iclr/HuSWALWWC22} modules to predict future attention patterns, thereby obtaining future-aware signals for KV selection.

Although these methods better align KV compression with generation
requirements, they generally introduce additional model components,
training procedures, or inference overhead, increasing system
complexity and deployment cost.

\section{Observations}
\label{sec:observations}

To characterize the transition from prompt-side observations to actual
decode-stage access patterns, we compare the attention distributions
produced by queries at different positions with a reference attention
distribution constructed from later decode queries. We consider both
distributional similarity and the overlap between their induced
KV-token selections.

Specifically, we evaluate the final $32$ prefill queries and the first
$8$ real decode queries, with relative positions
\begin{equation}
    r \in \{-32,-31,\ldots,-1,1,2,\ldots,8\}.
\end{equation}
Experiments are conducted using
Qwen3-8B~\cite{DBLP:journals/corr/abs-2505-09388},
Llama-3.1-8B-Instruct~\cite{DBLP:journals/corr/abs-2407-21783},
and
Mistral-7B-Instruct-v0.3~\cite{DBLP:journals/corr/abs-2310-06825}
on GovReport~\cite{DBLP:conf/naacl/HuangCPJW21}, MultiNews~\cite{DBLP:conf/acl/FabbriLSLR19}, QMSum~\cite{DBLP:conf/naacl/ZhongYYZMJACLQR21}, and RepoBench-P~\cite{DBLP:conf/iclr/0003XM24}.
For each model--task combination, we randomly sample $30$ examples,
resulting in $360$ evaluation instances in total. The KV budget is set
to $B=512$.

For each sample, we perform the complete generation process under the
FullKV setting. Let $N_n$ and $T_n$ denote the prompt length and the
number of generated tokens for sample $n$, respectively. To ensure a
fair comparison, all queries are evaluated over the same compressible
prompt prefix, which consists of the prompt positions preceding the
final $32$ tokens. For each query at relative position $r$, its
attention weights over this prefix are renormalized to form the
distribution $\mathbf{p}_{n,\ell,h}^{(r)}$ at layer $\ell$ and
attention head $h$.

Let $\mathbf{a}_{n,\ell,h}^{(t)}$ denote the normalized attention
distribution of the $t$-th decode query over the same prompt prefix.
We construct the reference attention distribution by averaging the
normalized attention distributions of the decode queries from step $9$
to the final generation step:
\begin{equation}
    \mathbf{o}_{n,\ell,h}
    =
    \frac{1}{T_n-8}
    \sum_{t=9}^{T_n}
    \mathbf{a}_{n,\ell,h}^{(t)}.
\end{equation}

\subsection{Metrics}
\label{sec:observation-metrics}

We evaluate the consistency between each query position and the reference
attention distribution from two perspectives.

\paragraph{JSD-Based Similarity.}
The Jensen--Shannon divergence (JSD)~\cite{DBLP:journals/tit/Lin91} is a symmetric measure of the
difference between two probability distributions. For the query
attention distribution $\mathbf{p}$ and the reference distribution
$\mathbf{o}$, let $\mathbf{m}=(\mathbf{p}+\mathbf{o})/2$. Their JSD is
defined as
\begin{equation}
\begin{aligned}
    \operatorname{JSD}(\mathbf{p},\mathbf{o})
    ={}&
    \frac{1}{2}
    D_{\mathrm{KL}}
    \left(
        \mathbf{p}\,\middle\|\,\mathbf{m}
    \right)
    +
    \frac{1}{2}
    D_{\mathrm{KL}}
    \left(
        \mathbf{o}\,\middle\|\,\mathbf{m}
    \right).
\end{aligned}
\end{equation}
With base-$2$ logarithms, JSD lies in $[0,1]$, where a smaller value
indicates greater similarity between the two attention distributions.
To facilitate trend comparison, we further define a JSD-based
similarity score as
\begin{equation}
    \operatorname{JSDSim}_{n,\ell,h}^{(r)}
    =
    1-
    \operatorname{JSD}
    \left(
        \mathbf{p}_{n,\ell,h}^{(r)},
        \mathbf{o}_{n,\ell,h}
    \right).
\end{equation}
A higher value indicates greater consistency between the query
attention distribution and the reference attention distribution.

\paragraph{KV Selection Recall.}
For sample $n$, layer $\ell$, and attention head $h$, let
$\mathcal{S}_{n,\ell,h}^{\star}$ denote the Top-$B$ prompt KV positions
selected according to the reference attention distribution, and let
$\mathcal{S}_{n,\ell,h}^{(r)}$ denote those selected according to the
attention scores of the query at relative position $r$. The corresponding
KV Selection Recall is defined as
\begin{equation}
    \operatorname{Recall}_{n,\ell,h}^{(r)}
    =
    \frac{
        \left|
        \mathcal{S}_{n,\ell,h}^{(r)}
        \cap
        \mathcal{S}_{n,\ell,h}^{\star}
        \right|
    }{B}.
\end{equation}

\begin{figure}[t]
    \centering
    \includegraphics[width=0.95\linewidth]
    {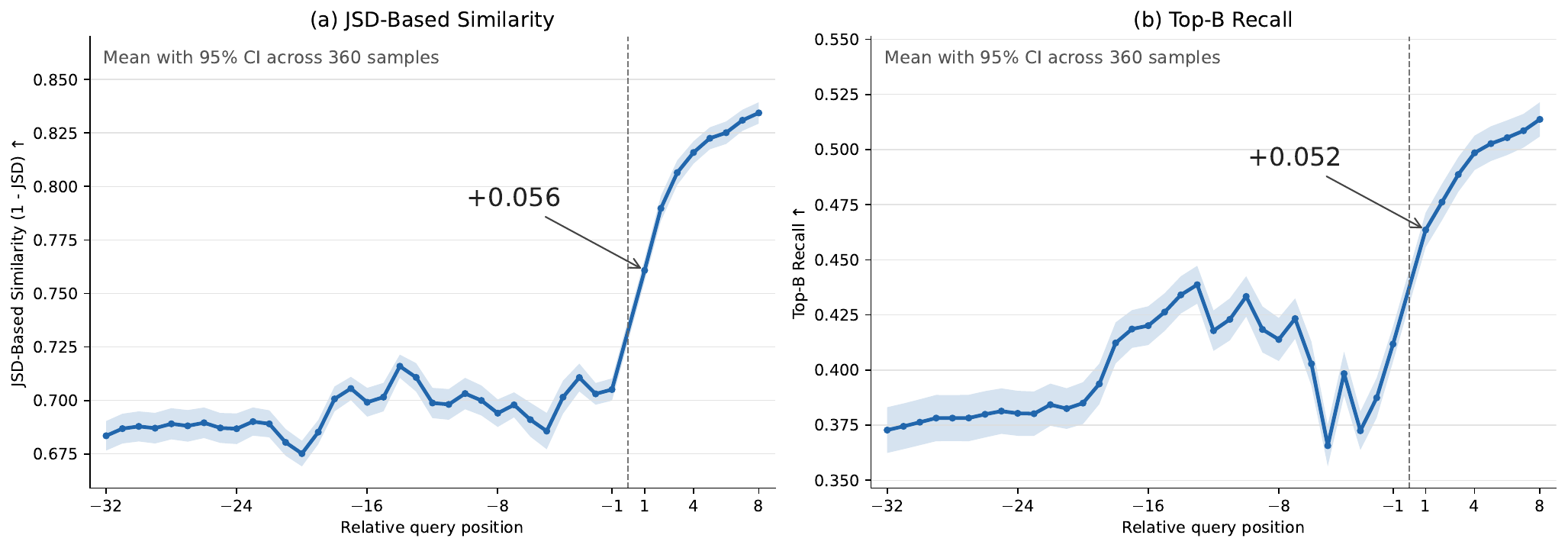}
    \caption{
        JSD-Based Similarity and Top-$B$ KV Selection Recall at different
        query positions.
    }
    \label{fig:query-position-jsd-recall}
\end{figure}

For each query position, we first average the metrics over all layers
and attention heads within each sample and then average them over all
$360$ evaluation instances. The shaded regions in
Figure~\ref{fig:query-position-jsd-recall} indicate $95\%$ confidence
intervals.

\subsection{Obs. 1: Prefill-to-Decode Attention Shift}
\label{sec:observation-prefill-decode-shift}

As shown in Figure~\ref{fig:query-position-jsd-recall}, prefill queries
already show moderate consistency with the reference attention
distribution, indicating that
prompt-side attention contains useful signals for KV importance
estimation~\cite{DBLP:conf/nips/LiHYVLYCLC24}. However, moving
from the final prefill query at $r=-1$ to the first real decode query at
$r=1$ increases JSD-Based Similarity and KV Selection Recall by $0.056$
and $0.052$, respectively.

This sharp improvement indicates that the first real decode query
reveals generation-specific access requirements that are not fully
captured during prefill, making it a more reliable signal for KV
importance estimation.

\subsection{Obs. 2: Largest Gain at First Decode}
\label{sec:observation-first-decode-query}

Among all adjacent query positions, the largest single-step gain occurs
at the prefill--decode boundary. In comparison, the average gains
between adjacent subsequent decode positions are only $0.009$ for
JSD-Based Similarity and $0.007$ for KV Selection Recall, both
substantially lower than those at the boundary.

Later decode queries can further improve consistency with the reference
attention distribution. However, using these later queries requires
retaining the full prompt KV cache for additional decode steps, thereby
increasing memory usage and end-to-end inference latency. Therefore,
the first real decode query provides the earliest decode-side observation
with the largest gain, offering a favorable trade-off between KV
selection quality and system overhead.

\begin{figure*}[t]
    \centering
    \includegraphics[width=0.95\textwidth]{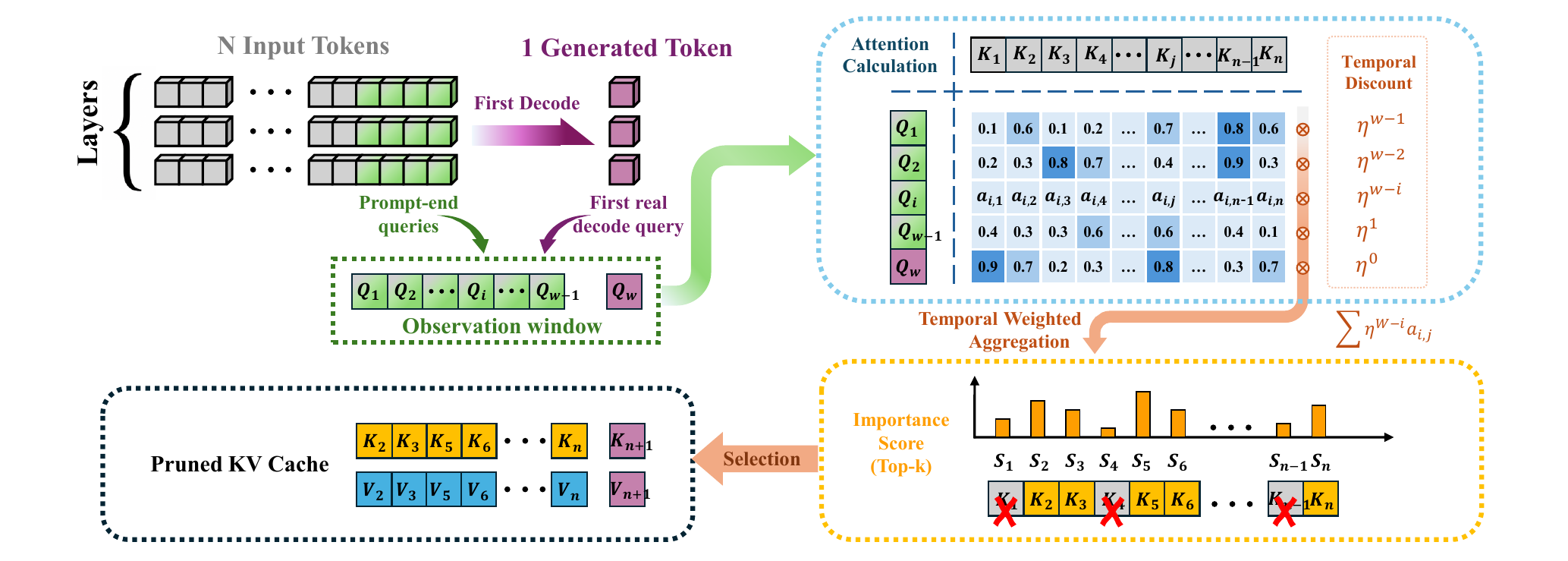}
    \caption{Overview of DeferKV. The first real decode query is combined with recent prompt queries and temporally weighted to estimate KV importance for subsequent cache compression.}
    \label{fig:framework}
\end{figure*}

\subsection{Obs. 3: Temporal Relevance Decay}
\label{sec:observation-temporal-decay}

Beyond the improvement at the prefill--decode boundary, both metrics
generally increase as query positions approach the decode stage.
Although the trend is not strictly monotonic, earlier prefill queries
are typically less consistent with the reference attention distribution.

This suggests that queries contribute differently to KV importance
estimation. We therefore assign larger weights to queries closer to the
decode stage and gradually discount earlier queries, emphasizing
generation-related signals while preserving prompt-side context.

\section{Method}
\label{sec:method}

\subsection{Overall Workflow}
\label{sec:method-overview}

Prompt-side attention does not fully reflect the KV access requirements
of subsequent decoding, while the first real decode query provides a
more reliable signal and queries closer to decoding are generally more
informative. Based on these observations, DeferKV postpones KV cache
compression until after the first standard decoding step.

As illustrated in Figure~\ref{fig:framework}, DeferKV obtains the first
real decode query from the target model and combines it with
recent prompt queries to form a decode-aware observation window. Their
attention scores are aggregated using temporal weighting, assigning
greater importance to queries closer to the decode stage. The resulting
scores are used to estimate prompt-token importance and construct the
compressed KV cache for subsequent generation. In this
way, DeferKV better aligns KV selection with actual decoding requirements
without predicting future queries, introducing auxiliary models, or
altering the standard generation trajectory.

% \begin{figure}[t]
%     \centering
%     \includegraphics[width=\columnwidth]
%     {Figures/try2.2.pdf}
%     \caption{Overview of DeferKV.}
%     \label{fig:framework}
% \end{figure}

\subsection{First Real Decode Query Integration}
\label{sec:decode-query-observation}

Let \(\mathbf{X}=\{x_1,x_2,\ldots,x_N\}\) denote a tokenized input prompt,
where $x_i$ denotes the token at position $i$ and $N$ denotes the prompt
length. The output logits at the final prompt position define
the conditional distribution of the first generated token. Accordingly,
the first token is obtained as
\begin{equation}
y_1
=
\operatorname{Decode}
\left(
p_{\theta}\left(\cdot \mid \mathbf{X}\right)
\right),
\end{equation}
where $p_{\theta}$ denotes the target model and
$\operatorname{Decode}(\cdot)$ denotes the adopted decoding strategy,
such as greedy decoding or sampling.

The generated token $y_1$ is then fed into the model for the first
standard decoding step. At layer $\ell$, the corresponding Query, Key,
and Value representations are denoted by
$\mathbf{q}_{\mathrm{dec},1}^{\ell}$,
$\mathbf{k}_{\mathrm{dec},1}^{\ell}$, and
$\mathbf{v}_{\mathrm{dec},1}^{\ell}$, respectively.

Here, $\mathbf{q}_{\mathrm{dec},1}^{\ell}$ is the real decode query produced by the target model when processing the first generated token. Given an observation window of size $W$, we combine this query with the last $W-1$ prompt queries to construct a new observation window:
\begin{equation}
\begin{aligned}
{\mathbf{Q}}_{\mathrm{obs}}^{\ell}
=
\big[
&\mathbf{q}_{N-W+2}^{\ell},
\ldots,
\mathbf{q}_{N}^{\ell}, \\
&\mathbf{q}_{\mathrm{dec},1}^{\ell}
\big].
\end{aligned}
\end{equation}

This design preserves the original observation-window size while incorporating information from the actual generation stage into KV-token importance estimation.

\subsection{Temporally Weighted Decode-Aware Eviction}
\label{sec:decode-aware-eviction}

Given the observation window defined above, we compute its attention weights over the prompt Keys as
\begin{equation}
{\mathbf{A}}^{\ell}
=
\operatorname{Attention}
\left(
{\mathbf{Q}}_{\mathrm{obs}}^{\ell},
\mathbf{K}^{\ell}
\right).
\end{equation} where $\mathbf{K}^{\ell}$ denotes the prompt Keys at the $\ell$-th layer. The attention computation preserves the original visibility constraints of the model. We denote the $(i,j)$-th element of ${\mathbf{A}}^{\ell}$ by ${a}_{i,j}^{\ell}$, representing the attention weight assigned by the $i$-th observation query to the $j$-th prompt Key.

The queries in the observation window are ordered chronologically, from
the earlier prompt queries to the first real decode query. Motivated by
the temporal relevance decay in Observation~3, we assign larger weights to
queries closer to the decode stage. The importance score of the $j$-th
prompt token is computed as
\begin{equation}
s_j^{\ell}
=
\sum_{i=1}^{W}
\eta^{W-i}
a_{i,j}^{\ell},
\qquad
0 < \eta \leq 1,
\end{equation}
where $\eta$ is the temporal discount factor. Therefore, the first real
decode query receives a weight of $1$, while earlier prompt queries are
exponentially discounted according to their temporal distance from the
decode stage. When $\eta=1$, this aggregation is equivalent to mean aggregation in terms of token ranking.

After obtaining the importance scores, we retain the local smoothing, budget allocation, and cache selection strategies of the underlying KV cache eviction method~\cite{DBLP:conf/nips/LiHYVLYCLC24}, while replacing its original prompt-only importance scores with the proposed decode-aware scores. The Key and Value corresponding to the first generated token are then combined with the compressed prompt KV cache to construct the KV cache used for subsequent decoding.

The model thereafter continues standard autoregressive generation. Our method only changes the timing of the initial KV cache compression and reuses the query naturally produced during the target model's decoding process. Therefore, it requires no modification to the model architecture or additional training procedure.

\section{Experiments}

\subsection{Experimental Setup}

% NOTE:
% The following values are synthetic placeholders for formatting only.
% Replace them with actual experimental results before submission.

\label{sec:experimental-setup}

\paragraph{Benchmarks and Models.}
We conduct experiments on LongBench~\cite{DBLP:conf/acl/BaiLZL0HDLZHDTL24},
RULER~\cite{DBLP:journals/corr/abs-2404-06654}, and Needle-in-a-Haystack~\cite{DBLP:journals/tacl/LiuLHPBPL24}. LongBench contains 16
long-context datasets across six task categories, covering question
answering, summarization, few-shot learning, synthetic tasks, and code
understanding. RULER evaluates model capabilities on
challenging long-context tasks across varying context lengths.
Needle-in-a-Haystack assesses the ability to retrieve critical information from long sequences, with detailed results reported in Appendix~\ref{app:niah}.

We evaluate Qwen3-8B~\cite{DBLP:journals/corr/abs-2505-09388},
Llama-3.1-8B-Instruct~\cite{DBLP:journals/corr/abs-2407-21783}, and Mistral-7B-Instruct-v0.3~\cite{DBLP:journals/corr/abs-2310-06825}. For each model, we use its official tokenizer and prompt template and adopt greedy decoding.

\begin{table*}[!t]
    \centering
    \setlength{\tabcolsep}{1.5pt}
    \renewcommand{\arraystretch}{0.98}
    \resizebox{\textwidth}{!}{%
    \begin{tabular}{@{}l*{17}{c}@{}}
        \toprule

        &
        \multicolumn{3}{c}{Single-Doc QA}
        &
        \multicolumn{3}{c}{Multi-Doc QA}
        &
        \multicolumn{3}{c}{Summ.}
        &
        \multicolumn{3}{c}{Few-Shot}
        &
        \multicolumn{2}{c}{Synth.}
        &
        \multicolumn{2}{c}{Code}
        &
        \\

        \cmidrule(lr){2-4}
        \cmidrule(lr){5-7}
        \cmidrule(lr){8-10}
        \cmidrule(lr){11-13}
        \cmidrule(lr){14-15}
        \cmidrule(lr){16-17}

        \textbf{Method}
        & \rot{NrtvQA}
        & \rot{Qasper}
        & \rot{MF-en}
        & \rot{HotpotQA}
        & \rot{2WikiMQA}
        & \rot{MuSiQue}
        & \rot{GovReport}
        & \rot{QMSum}
        & \rot{MultiNews}
        & \rot{TREC}
        & \rot{TriviaQA}
        & \rot{SAMSum}
        & \rot{PCount}
        & \rot{PR-en}
        & \rot{LCC}
        & \rot{RB-P}
        & \textbf{Avg.}
        \\

        % =========================================================
        % Qwen3-8B, Budget = 64L
        % =========================================================
        \midrule
        \multicolumn{18}{c}{
            \textbf{Qwen3-8B},
            \quad $B_{\mathrm{total}}=64L$
        }
        \\
        \midrule

        SnapKV & 17.41 & 30.06 & 38.05 & 47.77 & 37.71 & 20.65 & 14.00 & 19.37 & 12.93 & 43.50 & 81.92 & 36.53 & 5.00 & 87.50 & 59.30 & 53.58 & 37.83 \\
        LAQ & 18.99 & \textbf{37.88} & 43.31 & \underline{53.88} & \underline{41.35} & 23.24 & \textbf{16.18} & 19.40 & \underline{14.78} & \underline{66.00} & \underline{88.40} & \textbf{37.64} & 5.50 & 94.50 & \underline{62.31} & \underline{55.91} & \underline{42.45} \\
        SpecKV & \underline{19.61} & 29.97 & \textbf{49.79} & 51.03 & 36.97 & 22.14 & 13.27 & \textbf{19.72} & \textbf{14.82} & 37.50 & 73.63 & 36.71 & \underline{6.00} & \underline{95.00} & 45.15 & 46.62 & 37.37 \\
        LookaheadKV & 18.22 & 32.27 & 40.12 & 47.66 & 38.01 & \underline{23.58} & 14.61 & 18.36 & 13.18 & 46.95 & 83.29 & 36.69 & 5.00 & 90.40 & 59.06 & 54.28 & 38.86 \\
        \textbf{DeferKV} & \textbf{21.87} & \underline{35.37} & \underline{45.97} & \textbf{56.99} & \textbf{41.88} & \textbf{31.20} & \underline{14.81} & \underline{19.52} & 13.52 & \textbf{70.50} & \textbf{89.98} & \underline{37.44} & \textbf{8.00} & \textbf{97.50} & \textbf{62.52} & \textbf{56.47} & \textbf{43.97} \\

        % =========================================================
        % Qwen3-8B, Budget = 512L
        % =========================================================
        \midrule
        \multicolumn{18}{c}{
            \textbf{Qwen3-8B},
            \quad $B_{\mathrm{total}}=512L$
        }
        \\
        \midrule

        SnapKV & 22.70 & 42.96 & 51.19 & 57.09 & 42.51 & \textbf{34.94} & 25.14 & \textbf{23.44} & 21.87 & 68.50 & 89.46 & 43.34 & \textbf{9.50} & \textbf{100.00} & \textbf{68.74} & 64.22 & 47.85 \\
        LAQ & 23.16 & 43.58 & \underline{52.63} & 56.16 & 43.02 & 34.36 & 25.46 & 22.02 & 21.94 & \underline{71.00} & \textbf{90.05} & 43.35 & \textbf{9.50} & \textbf{100.00} & 67.92 & \underline{64.72} & 48.05 \\
        SpecKV & \underline{24.38} & 43.27 & \textbf{53.93} & \underline{58.07} & 42.90 & 31.48 & \textbf{29.51} & \underline{23.24} & \textbf{24.05} & \underline{71.00} & 89.96 & 29.60 & \underline{4.00} & \underline{99.00} & 56.84 & 56.85 & 46.13 \\
        LookaheadKV & 23.10 & \underline{43.67} & 52.21 & 56.86 & \underline{43.12} & \underline{34.91} & 25.46 & 22.71 & 22.08 & \underline{71.00} & \underline{89.97} & \textbf{43.85} & \textbf{9.50} & \underline{99.00} & \underline{68.01} & \textbf{64.98} & \underline{48.15} \\
        \textbf{DeferKV} & \textbf{24.82} & \textbf{44.78} & 51.96 & \textbf{58.09} & \textbf{43.18} & 33.83 & \underline{26.04} & 22.74 & \underline{22.17} & \textbf{71.50} & \textbf{90.05} & \underline{43.70} & \textbf{9.50} & \textbf{100.00} & 67.55 & 64.02 & \textbf{48.37} \\

        \midrule
        Full KV & 24.73 & 48.05 & 54.39 & 58.16 & 43.79 & 34.27 & 33.45 & 24.12 & 24.84 & 71.50 & 90.21 & 44.56 & 9.50 & 100.00 & 69.34 & 65.48 & 49.77 \\

        % =========================================================
        % Llama-3.1-8B-Instruct, Budget = 64L
        % =========================================================
        \midrule
        \multicolumn{18}{c}{
            \textbf{Llama-3.1-8B-Instruct},
            \quad $B_{\mathrm{total}}=64L$
        }
        \\
        \midrule

        SnapKV & 27.32 & 25.25 & 44.30 & 53.98 & 42.48 & 27.10 & 17.88 & 22.20 & 17.45 & 41.00 & 85.37 & 37.72 & \underline{8.50} & \textbf{100.00} & 55.35 & 48.69 & 40.91 \\
        LAQ & 27.66 & \underline{28.03} & 47.05 & \textbf{58.12} & \underline{45.06} & \underline{30.60} & 18.85 & 22.51 & \underline{19.01} & \textbf{71.50} & \textbf{90.02} & 40.16 & \underline{8.50} & \underline{99.50} & 56.39 & \underline{51.29} & \underline{44.64} \\
        SpecKV & \underline{29.31} & 23.34 & \textbf{56.27} & 56.81 & 41.91 & 28.70 & \textbf{21.22} & \textbf{24.43} & 18.66 & 46.98 & 76.62 & 39.69 & \textbf{9.71} & \textbf{100.00} & 41.67 & 42.44 & 41.11 \\
        LookaheadKV & 27.95 & 26.61 & 47.23 & 54.93 & 44.90 & 30.10 & 19.24 & 22.01 & 18.34 & \underline{70.00} & 88.65 & \underline{40.21} & \underline{8.50} & 99.00 & \underline{56.54} & 49.03 & 43.95 \\
        \textbf{DeferKV} & \textbf{29.83} & \textbf{31.02} & \underline{49.95} & \underline{57.55} & \textbf{47.65} & \textbf{30.77} & \underline{20.68} & \underline{22.77} & \textbf{19.29} & \textbf{71.50} & \underline{88.95} & \textbf{40.41} & 7.88 & \textbf{100.00} & \textbf{57.98} & \textbf{51.56} & \textbf{45.49} \\

        % =========================================================
        % Llama-3.1-8B-Instruct, Budget = 512L
        % =========================================================
        \midrule
        \multicolumn{18}{c}{
            \textbf{Llama-3.1-8B-Instruct},
            \quad $B_{\mathrm{total}}=512L$
        }
        \\
        \midrule

        SnapKV & 29.92 & 40.20 & 55.21 & 57.85 & 48.02 & 30.03 & 25.79 & \underline{24.40} & 24.33 & 65.00 & \textbf{91.95} & 41.80 & \textbf{8.08} & \underline{99.50} & \textbf{64.39} & \underline{57.27} & 47.73 \\
        LAQ & \underline{30.58} & \underline{42.99} & 54.46 & 58.13 & 48.40 & 30.79 & 26.45 & 24.15 & 24.51 & \underline{70.96} & 91.51 & \underline{42.74} & 7.59 & \underline{99.50} & \underline{64.08} & 56.78 & \underline{48.35} \\
        SpecKV & 28.23 & 41.70 & \textbf{56.69} & \textbf{59.29} & \underline{48.62} & 28.24 & \textbf{30.41} & \textbf{26.44} & \textbf{27.08} & \textbf{72.00} & 91.63 & 28.51 & 3.23 & 99.00 & 53.07 & 49.99 & 46.51 \\
        LookaheadKV & 30.11 & 41.12 & \underline{55.34} & 57.98 & \textbf{49.23} & \textbf{31.34} & 26.02 & 23.85 & 24.06 & \textbf{72.00} & \underline{91.85} & 41.64 & \underline{7.75} & \underline{99.50} & 63.21 & \textbf{57.33} & 48.27 \\
        \textbf{DeferKV} & \textbf{30.70} & \textbf{43.91} & 54.84 & \underline{58.54} & 48.51 & \underline{31.02} & \underline{26.60} & 24.22 & \underline{24.53} & \textbf{72.00} & 91.48 & \textbf{42.98} & 7.47 & \textbf{100.00} & 63.90 & 56.70 & \textbf{48.59} \\

        \midrule
        Full KV & 31.20 & 46.66 & 56.64 & 58.10 & 48.51 & 31.57 & 34.37 & 25.18 & 27.14 & 73.00 & 91.65 & 43.80 & 7.64 & 99.50 & 65.18 & 58.90 & 49.94 \\

        % =========================================================
        % Mistral-7B-Instruct-v0.3, Budget = 64L
        % =========================================================
        \midrule
        \multicolumn{18}{c}{
            \textbf{Mistral-7B-Instruct-v0.3},
            \quad $B_{\mathrm{total}}=64L$
        }
        \\
        \midrule

        SnapKV & 17.65 & 22.08 & 36.24 & 41.95 & 33.98 & 20.89 & 18.83 & 21.14 & 16.34 & 38.50 & 86.96 & 39.81 & 4.50 & 77.00 & 51.57 & 49.59 & 36.06 \\
        LAQ & 19.72 & \underline{28.68} & 40.14 & \underline{48.46} & \underline{35.96} & \underline{23.44} & \underline{20.82} & 21.47 & \textbf{18.93} & \underline{70.58} & \textbf{89.15} & \textbf{42.32} & 4.75 & 79.80 & \underline{53.08} & \underline{51.19} & \underline{40.53} \\
        SpecKV & \underline{19.74} & 21.08 & \textbf{48.10} & 45.00 & 33.32 & 22.44 & \textbf{22.64} & \textbf{22.93} & 17.89 & 38.49 & 78.44 & 40.89 & \underline{5.60} & \underline{80.75} & 37.66 & 42.98 & 36.12 \\
        LookaheadKV & 19.48 & 24.28 & 38.88 & 42.86 & 34.76 & 23.22 & \underline{20.82} & 21.02 & \textbf{18.93} & 57.38 & 88.09 & 41.11 & 4.50 & 78.16 & 51.92 & 49.93 & 38.46 \\
        \textbf{DeferKV} & \textbf{26.11} & \textbf{28.83} & \underline{43.02} & \textbf{49.10} & \textbf{36.87} & \textbf{25.15} & 20.49 & \underline{21.75} & \underline{18.50} & \textbf{73.50} & \underline{89.04} & \underline{42.17} & \textbf{7.50} & \textbf{81.00} & \textbf{53.41} & \textbf{51.46} & \textbf{41.74} \\

        % =========================================================
        % Mistral-7B-Instruct-v0.3, Budget = 512L
        % =========================================================
        \midrule
        \multicolumn{18}{c}{
            \textbf{Mistral-7B-Instruct-v0.3},
            \quad $B_{\mathrm{total}}=512L$
        }
        \\
        \midrule

        SnapKV & 25.48 & 33.15 & \underline{49.37} & 48.98 & 36.54 & 24.68 & 25.61 & \underline{24.27} & 24.11 & 67.50 & \underline{89.36} & 44.82 & \textbf{6.00} & \textbf{97.50} & \textbf{60.36} & \underline{61.28} & 44.94 \\
        LAQ & \underline{26.23} & \underline{34.50} & 48.82 & 49.95 & 38.22 & 26.58 & 26.10 & 24.10 & 24.20 & 74.66 & 89.08 & \underline{47.10} & \textbf{6.00} & \textbf{97.50} & \underline{60.02} & \textbf{61.44} & \underline{45.91} \\
        SpecKV & 25.48 & 34.10 & \textbf{51.48} & \underline{50.19} & 37.03 & 22.06 & \textbf{30.10} & \textbf{25.69} & \textbf{25.58} & 72.25 & \textbf{89.45} & 31.30 & \underline{3.83} & \underline{96.75} & 48.75 & 53.96 & 43.63 \\
        LookaheadKV & 25.79 & 33.86 & 49.15 & 48.94 & \textbf{39.16} & \underline{26.67} & 25.88 & 24.12 & 23.93 & \underline{75.29} & 89.20 & 46.33 & \textbf{6.00} & \textbf{97.50} & 59.54 & \textbf{61.44} & 45.80 \\
        \textbf{DeferKV} & \textbf{26.90} & \textbf{35.37} & 48.91 & \textbf{51.10} & \underline{38.72} & \textbf{27.63} & \underline{26.44} & 24.13 & \underline{24.27} & \textbf{76.00} & 89.07 & \textbf{47.18} & \textbf{6.00} & \textbf{97.50} & 59.68 & 60.49 & \textbf{46.21} \\

        \midrule
        Full KV & 27.01 & 37.97 & 50.14 & 51.69 & 39.17 & 28.02 & 34.27 & 25.66 & 26.62 & 76.00 & 88.59 & 47.50 & 7.00 & 98.00 & 61.53 & 62.59 & 47.61 \\

        \bottomrule
    \end{tabular}%
    }

    \caption{Performance comparison on 16 LongBench datasets.
    The best and second-best results among the five cache-compression methods
    are highlighted in bold and underlined, respectively.
    Full KV is included only as a reference.}
    \label{tab:main-results}
\end{table*}

\label{sec:main-results}
\paragraph{Selected Baselines.}
We compare against four representative KV cache compression methods:
(1) \textbf{SnapKV}~\cite{DBLP:conf/nips/LiHYVLYCLC24}, which estimates historical KV importance using a
local observation window near the end of the prompt and performs
one-time compression before decoding; (2) \textbf{LAQ}~\cite{DBLP:conf/emnlp/WangJLXXZC25}, which performs low-budget lookahead decoding with an initially compressed cache and uses the resulting query states to guide a second KV eviction; (3) \textbf{SpecKV}~\cite{galim2026draftbased}, which uses a small draft model to generate a proxy response and processes it with the target model to estimate prompt-KV importance; (4) \textbf{LookaheadKV}~\cite{DBLP:journals/corr/abs-2603-10899}, which uses learnable lookahead tokens and lightweight LoRA modules to estimate prompt-KV importance.

\paragraph{Implementation Details.}
We evaluate DeferKV and all baselines under varying cache budgets
($B_{\mathrm{total}} = nL$, with $n \in [64, 1024]$), where $n$ is the prompt KV budget for each layer and $L$ is the number of layers. We
use greedy decoding without sampling for all experiments to ensure
deterministic and reproducible evaluation. We use the official baseline implementations with standard eviction settings, with further details provided in Appendix~\ref{app:implementation-details}. All experiments are conducted
on NVIDIA A100 GPUs with 80\,GB of memory.

\subsection{Results on LongBench}

We evaluate DeferKV on 16 LongBench datasets using three language
models. Table~\ref{tab:main-results} reports detailed results under a tight cache
budget of $B_{\mathrm{total}}=64L$ and a larger cache budget of
$B_{\mathrm{total}}=512L$. DeferKV achieves the highest average score
across all six model--budget settings.

Under the tight $64L$ budget, DeferKV consistently ranks as the
top-performing method across models, achieving up to a 3.58\% relative
improvement over the strongest baseline with Qwen3-8B. This advantage is particularly pronounced under constrained cache
budgets, where real decode feedback improves KV importance estimation
and helps preserve information essential for subsequent generation.

As the budget increases to $B_{\mathrm{total}}=512L$, the performance
gaps among methods narrow because more prompt KV tokens can be retained.
Nevertheless, DeferKV continues to achieve the best average performance
and approaches Full KV across all three models. In addition to its strong
performance, DeferKV requires no additional training, draft model, or
future-query prediction module, making it simple to implement and deploy.
We further integrate DeferKV with PyramidKV, AdaKV, and ChunkKV to
demonstrate its compatibility with existing KV cache compression methods.
The corresponding results are provided in Appendix~\ref{app:compatibility}.

\subsection{Results on RULER}

\begin{table}[t]
\centering
\small
\setlength{\tabcolsep}{1.5pt}
\begin{tabular*}{\columnwidth}{
    @{\extracolsep{\fill}}
    l r r r r r r r
    @{}
}
\toprule
\textbf{Method}
& \textbf{4K}
& \textbf{8K}
& \textbf{16K}
& \textbf{32K}
& \textbf{64K}
& \textbf{128K}
& \textbf{Avg.} \\
\midrule
Full KV       & 99.34 & 98.83 & 98.55 & 94.89 & 89.85 & 79.32 & 93.46 \\
\midrule
SnapKV        & 83.63 & 75.51 & 71.09 & 66.12 & 56.26 & 45.67 & 66.38 \\
LAQ           & 87.42 & 81.78 & 78.27 & 74.09 & 68.23 & 62.06 & 75.31 \\
SpecKV        & 86.14 & 79.43 & 75.60 & 70.53 & 62.83 & 54.17 & 71.45 \\
LookaheadKV   & 86.66 & 80.69 & 76.98 & 72.43 & 65.73 & 59.26 & 73.63 \\
\textbf{DeferKV}
              & \textbf{92.72}
              & \textbf{86.29}
              & \textbf{83.89}
              & \textbf{80.31}
              & \textbf{75.89}
              & \textbf{69.04}
              & \textbf{81.36} \\
\bottomrule
\end{tabular*}
\caption{Performance comparison on the RULER benchmark across different
context lengths.}
\label{tab:ruler_results}
\end{table}

We evaluate DeferKV on 11 tasks from the RULER benchmark using
Llama-3.1-8B-Instruct, with a fixed total cache budget of
$B_{\mathrm{total}}=1024L$ applied across all methods.
Table~\ref{tab:ruler_results} summarizes the average accuracy across
context lengths ranging from 4K to 128K. Existing KV cache eviction
methods suffer from substantial performance degradation as the context
length increases, highlighting their limited robustness in
ultra-long-context and complex retrieval scenarios.

In contrast, DeferKV consistently achieves strong performance across all
context lengths, improving the average accuracy by 8.03\% over the
strongest baseline. Notably, even at a context length of
128K, where less than 1\% of the original KV cache is retained, DeferKV
maintains effective retrieval capability. These results demonstrate that incorporating a real decode query provides a more effective generation-aware signal for selecting critical KV tokens in ultra-long-context scenarios. 

% \subsection{Results on Needle-in-a-Haystack}

% \begin{figure}[t]
%     \centering
%     \includegraphics[
%         width=0.9\linewidth
%     ]{Figures/Mistral7B_SnapKV_b358_vs_DeferKV_b128_800_32000.pdf}
%     \caption{
%         Needle-in-a-Haystack performance of LookaheadKV and DeferKV.
%     }
%     \label{fig:needle-results}
% \end{figure}

% To further evaluate the effectiveness and robustness of DeferKV, we
% conduct experiments on the Needle-in-a-Haystack
% test using Mistral-7B-Instruct-v0.3 with a total KV
% cache budget of $B_{\mathrm{total}}=128L$. As shown in
% Figure~\ref{fig:needle-results}, even under such a constrained cache
% budget, DeferKV achieves a mean retrieval score of $0.892$ and outperforms LookaheadKV. This further demonstrates that
% DeferKV can effectively prioritize and retain the KV pairs most relevant
% to subsequent retrieval.

\subsection{Efficiency Analysis}
\label{sec:efficiency-analysis}
\paragraph{Peak Memory and Inference Latency.}
We evaluate inference efficiency using peak GPU memory and total
end-to-end latency. All methods use Qwen3-8B with a 32K-token input, 128 generated tokens,
and a cache budget of $B_{\mathrm{total}}=1024L$. We report the average results over five runs after warm-up.

\begin{figure}[t]
    \centering
    \includegraphics[
        width=0.9\linewidth
    ]{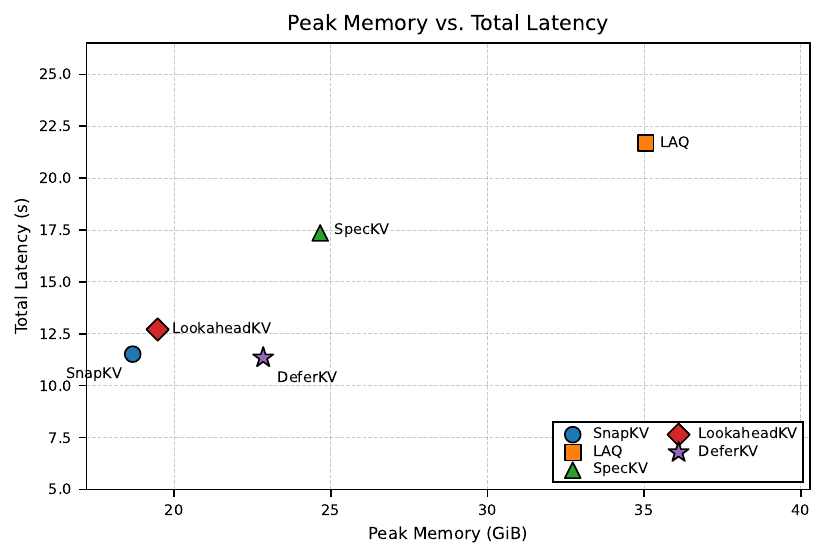}
    \caption{
        Peak GPU memory and end-to-end inference latency of different
    methods on Qwen3-8B.
    }
    \label{fig:efficiency-comparison}
\end{figure}

Figure~\ref{fig:efficiency-comparison} compares the peak GPU memory and
end-to-end inference latency of different methods. DeferKV achieves a
latency of $11.56$\,s, comparable to SnapKV and lower than LAQ, SpecKV,
and LookaheadKV. Its peak memory usage is $22.84$\,GiB, lower than LAQ
and SpecKV, although higher than SnapKV and LookaheadKV. Overall,
DeferKV provides a favorable balance between memory usage and inference
latency.

\paragraph{Effect of Compression Delay Steps.}

\begin{table}[t]
\centering
\small

\begin{tabular}{c r r r}
\toprule
\textbf{Delay $d$}
& \textbf{Score} $\uparrow$
& \textbf{Peak Mem.} $\downarrow$
& \textbf{Latency} $\downarrow$
\\
\midrule
$\mathbf{1}$ & 27.06 & \textbf{22.84} & \textbf{11.562} \\
$2$          & 27.11 & 22.85          & 11.618          \\
$3$          & 27.15 & 22.85          & 11.671          \\
$4$          & 27.21 & 22.86          & 11.751          \\
\bottomrule
\end{tabular}

\caption{Effect of compression delay steps on 32K-token GovReport samples using Qwen3-8B. Peak memory is reported in GiB, and latency
is reported in seconds.}
\label{tab:delay-step-efficiency}
\end{table}

We vary the number of real decode queries observed before compression
as $d\in\{1,2,3,4\}$, where $d=1$ is the default DeferKV setting.
Larger values retain the full prompt KV cache for additional decoding
steps before compression. All settings are evaluated on 100 randomly selected 32K-token samples from GovReport using Qwen3-8B, with 128 generated tokens and a cache budget of $B_{\mathrm{total}}=1024L$. Task scores are averaged over the 100 samples, while memory and latency are measured on a single sample and averaged over five runs after warm-up.

As shown in Table~\ref{tab:delay-step-efficiency}, delaying compression
for additional decode steps yields only marginal performance gains.
Larger $d$ values keep the full prompt KV cache resident for longer and
require more decoding steps before eviction, leading to increased
end-to-end latency. We therefore adopt $d=1$ as the default setting.

\subsection{Ablation Study}

We conduct ablation studies on LongBench to evaluate the contributions
of real decode feedback and temporally weighted aggregation. We use
Qwen3-8B with a cache budget of
$B_{\mathrm{total}}=128L$.

\begin{table}[t]
\centering
\small

\begin{tabular}{l r}
\toprule
\textbf{Method} & \textbf{Avg. Score} $\uparrow$ \\
\midrule
DeferKV
    & \textbf{46.12} \\
w/o Decode Feedback
    & 44.38 ($-$1.74) \\
w/o Temporal Weighting
    & 45.45 ($-$0.67) \\
\bottomrule
\end{tabular}

\caption{Ablation results of DeferKV on LongBench using Qwen3-8B.}
\label{tab:ablation-study}
\end{table}

As shown in Table~\ref{tab:ablation-study}, removing real decode
feedback decreases the average score by $1.74$, showing that the first
real decode query provides useful information for KV importance
estimation. Replacing temporal weighting with uniform aggregation
reduces the score by $0.67$, confirming the benefit of emphasizing
queries closer to the decode stage. Both components contribute to the
effectiveness of DeferKV.

\begin{figure}[!htbp]
    \centering
    \includegraphics[
        width=0.9\linewidth
    ]{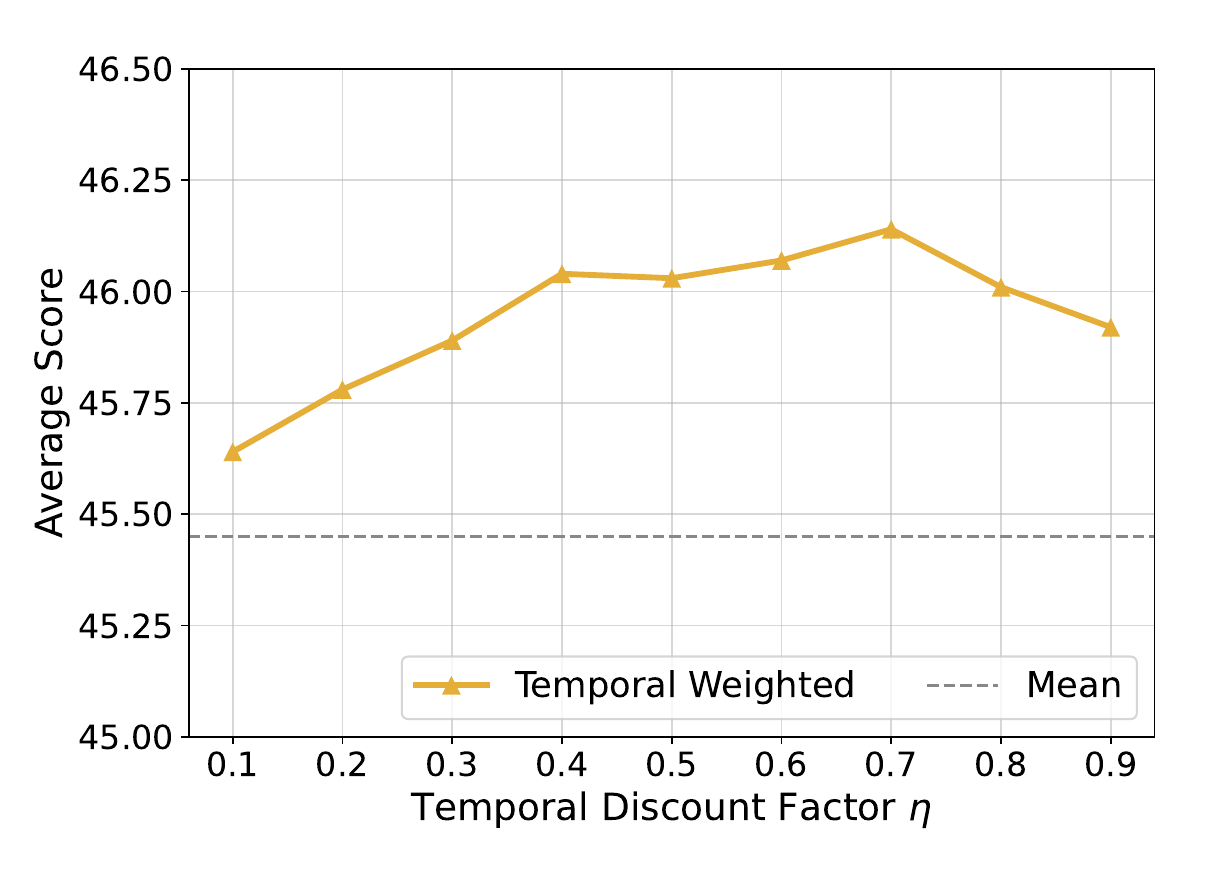}
    \caption{
        Effect of the temporal discount factor $\eta$ on model
        performance. 
    }
    \label{fig:eta-analysis}
\end{figure}

\subsection{Hyperparameter Sensitivity}

We evaluate the effect of the temporal discount factor $\eta$ on
LongBench using Qwen3-8B with a cache budget of
$B_{\mathrm{total}}=128L$. As shown in
Figure~\ref{fig:eta-analysis}, DeferKV maintains stable performance
over a wide range of values. Based on these results, we uniformly set
$\eta=0.7$ for all models and tasks without additional tuning for
specific scenarios.

\section{Conclusion} 

In this paper, we revisit a basic design choice in one-shot KV cache compression: when to commit to irreversible eviction. DeferKV moves this decision to the first decoding step, where real generation-side feedback becomes available, and uses temporal weighting to integrate the first real decode query with recent prompt queries. Extensive evaluations on LongBench, Needle-in-a-Haystack, and RULER show that DeferKV improves long-context performance with low inference latency. It requires no additional training, draft model, or auxiliary prediction module, providing a lightweight approach to efficient long-context inference.

\section*{Limitations}

DeferKV postpones one-shot KV eviction until the earliest real decode query becomes available, which requires retaining the full prompt KV cache through the first decoding step. This design therefore shifts the memory reduction to subsequent decoding. Our experiments focus on models with standard token-wise KV caches under softmax attention. Extending DeferKV to architectures with different cache organizations, such as MLA or hybrid linear-attention models, may require architecture-specific adaptation.

\bibliography{custom}

@inproceedings{DBLP:conf/nips/LiHYVLYCLC24,
  author       = {Yuhong Li and
                  Yingbing Huang and
                  Bowen Yang and
                  Bharat Venkitesh and
                  Acyr Locatelli and
                  Hanchen Ye and
                  Tianle Cai and
                  Patrick Lewis and
                  Deming Chen},
  title        = {SnapKV: {LLM} Knows What You are Looking for Before Generation},
  booktitle    = {NeurIPS},
  year         = {2024}
}

@inproceedings{galim2026draftbased,
title={Draft-based Approximate Inference for {LLM}s},
author={Kevin Galim and Ethan Ewer and Wonjun Kang and Minjae Lee and Hyung Il Koo and Kangwook Lee},
booktitle={The Fourteenth International Conference on Learning Representations},
year={2026}
}

@inproceedings{DBLP:conf/emnlp/WangJLXXZC25,
  author       = {Yixuan Wang and
                  Shiyu Ji and
                  Yijun Liu and
                  Yuzhuang Xu and
                  Yang Xu and
                  Qingfu Zhu and
                  Wanxiang Che},
  title        = {Lookahead Q-Cache: Achieving More Consistent {KV} Cache Eviction via
                  Pseudo Query},
  booktitle    = {EMNLP},
  pages        = {34158--34174},
  year         = {2025}
}

@article{DBLP:journals/corr/abs-2603-10899,
  author       = {Jinwoo Ahn and
                  Ingyu Seong and
                  Akhil Kedia and
                  Junhan Kim and
                  Hyemi Jang and
                  Kangwook Lee and
                  Yongkweon Jeon},
  title        = {LookaheadKV: Fast and Accurate {KV} Cache Eviction by Glimpsing into
                  the Future without Generation},
  journal      = {CoRR},
  volume       = {abs/2603.10899},
  year         = {2026},
}

@article{DBLP:journals/corr/abs-2406-02069,
  author       = {Zefan Cai and
                  Yichi Zhang and
                  Bofei Gao and
                  Yuliang Liu and
                  Tianyu Liu and
                  Keming Lu and
                  Wayne Xiong and
                  Yue Dong and
                  Baobao Chang and
                  Junjie Hu and
                  Wen Xiao},
  title        = {PyramidKV: Dynamic {KV} Cache Compression based on Pyramidal Information
                  Funneling},
  journal      = {CoRR},
  volume       = {abs/2406.02069},
  year         = {2024},
}

@inproceedings{DBLP:conf/nips/FengLCXZ25,
  author       = {Yuan Feng and
                  Junlin Lv and
                  Yukun Cao and
                  Xike Xie and
                  S. Kevin Zhou},
  title        = {Ada-KV: Optimizing {KV} Cache Eviction by Adaptive Budget Allocation
                  for Efficient {LLM} Inference},
  booktitle    = {NeurIPS},
  year         = {2025}
}

@inproceedings{DBLP:conf/acl/BaiLZL0HDLZHDTL24,
  author       = {Yushi Bai and
                  Xin Lv and
                  Jiajie Zhang and
                  Hongchang Lyu and
                  Jiankai Tang and
                  Zhidian Huang and
                  Zhengxiao Du and
                  Xiao Liu and
                  Aohan Zeng and
                  Lei Hou and
                  Yuxiao Dong and
                  Jie Tang and
                  Juanzi Li},
  title        = {LongBench: {A} Bilingual, Multitask Benchmark for Long Context Understanding},
  booktitle    = {ACL},
  pages        = {3119--3137},
  year         = {2024},
}

@article{DBLP:journals/corr/abs-2505-09388,
  author       = {Qwen Team},
  title        = {Qwen3 Technical Report},
  journal      = {CoRR},
  volume       = {abs/2505.09388},
  year         = {2025},
}

@article{DBLP:journals/corr/abs-2407-21783,
  author       = {Llama Team},
  title        = {The Llama 3 Herd of Models},
  journal      = {CoRR},
  volume       = {abs/2407.21783},
  year         = {2024},
}

@article{DBLP:journals/tmlr/0002LTTXCHD0025,
  author       = {Haoyang Li and
                  Yiming Li and
                  Anxin Tian and
                  Tianhao Tang and
                  Zhanchao Xu and
                  Xuejia Chen and
                  Nicole Hu and
                  Wei Dong and
                  Qing Li and
                  Lei Chen},
  title        = {A Survey on Large Language Model Acceleration based on {KV} Cache
                  Management},
  journal      = {Trans. Mach. Learn. Res.},
  volume       = {2025},
  year         = {2025}
}

@article{DBLP:journals/corr/abs-2310-06825,
  author       = {Albert Q. Jiang and
                  Alexandre Sablayrolles and
                  Arthur Mensch and
                  Chris Bamford and
                  Devendra Singh Chaplot and
                  Diego de Las Casas and
                  Florian Bressand and
                  Gianna Lengyel and
                  Guillaume Lample and
                  Lucile Saulnier and
                  L{\'{e}}lio Renard Lavaud and
                  Marie{-}Anne Lachaux and
                  Pierre Stock and
                  Teven Le Scao and
                  Thibaut Lavril and
                  Thomas Wang and
                  Timoth{\'{e}}e Lacroix and
                  William El Sayed},
  title        = {Mistral 7B},
  journal      = {CoRR},
  volume       = {abs/2310.06825},
  year         = {2023}
}

@article{DBLP:journals/tit/Lin91,
  author       = {Jianhua Lin},
  title        = {Divergence measures based on the Shannon entropy},
  journal      = {{IEEE} Trans. Inf. Theory},
  volume       = {37},
  number       = {1},
  pages        = {145--151},
  year         = {1991}
}

@article{DBLP:journals/corr/abs-2404-06654,
  author       = {Cheng{-}Ping Hsieh and
                  Simeng Sun and
                  Samuel Kriman and
                  Shantanu Acharya and
                  Dima Rekesh and
                  Fei Jia and
                  Yang Zhang and
                  Boris Ginsburg},
  title        = {{RULER:} What's the Real Context Size of Your Long-Context Language
                  Models?},
  journal      = {CoRR},
  volume       = {abs/2404.06654},
  year         = {2024}
}

@article{DBLP:journals/corr/abs-2502-00299,
  author       = {Xiang Liu and
                  Zhenheng Tang and
                  Peijie Dong and
                  Zeyu Li and
                  Bo Li and
                  Xuming Hu and
                  Xiaowen Chu},
  title        = {ChunkKV: Semantic-Preserving {KV} Cache Compression for Efficient
                  Long-Context {LLM} Inference},
  journal      = {CoRR},
  volume       = {abs/2502.00299},
  year         = {2025}
}

@article{DBLP:journals/corr/abs-2407-18003,
  author       = {Luohe Shi and
                  Hongyi Zhang and
                  Yao Yao and
                  Zuchao Li and
                  Hai Zhao},
  title        = {Keep the Cost Down: {A} Review on Methods to Optimize LLM' s
                  KV-Cache Consumption},
  journal      = {CoRR},
  volume       = {abs/2407.18003},
  year         = {2024}
}

@inproceedings{DBLP:conf/emnlp/YuanLZCLWLJCX0H24,
  author       = {Jiayi Yuan and
                  Hongyi Liu and
                  Shaochen (Henry) Zhong and
                  Yu{-}Neng Chuang and
                  Songchen Li and
                  Guanchu Wang and
                  Duy Le and
                  Hongye Jin and
                  Vipin Chaudhary and
                  Zhaozhuo Xu and
                  Zirui Liu and
                  Xia Ben Hu},
  title        = {{KV} Cache Compression, But What Must We Give in Return? {A} Comprehensive
                  Benchmark of Long Context Capable Approaches},
  booktitle    = {EMNLP},
  volume       = {{EMNLP} 2024},
  pages        = {4623--4648},
  year         = {2024}
}

@article{DBLP:journals/corr/abs-2303-08774,
  author       = {OpenAI},
  title        = {{GPT-4} Technical Report},
  journal      = {CoRR},
  volume       = {abs/2303.08774},
  year         = {2023}
}

@inproceedings{DBLP:conf/iclr/HuSWALWWC22,
  author       = {Edward J. Hu and
                  Yelong Shen and
                  Phillip Wallis and
                  Zeyuan Allen{-}Zhu and
                  Yuanzhi Li and
                  Shean Wang and
                  Lu Wang and
                  Weizhu Chen},
  title        = {LoRA: Low-Rank Adaptation of Large Language Models},
  booktitle    = {ICLR},
  year         = {2022}
}

@inproceedings{DBLP:conf/acl/FabbriLSLR19,
  author       = {Alexander R. Fabbri and
                  Irene Li and
                  Tianwei She and
                  Suyi Li and
                  Dragomir R. Radev},
  title        = {Multi-News: {A} Large-Scale Multi-Document Summarization Dataset and
                  Abstractive Hierarchical Model},
  booktitle    = {ACL},
  pages        = {1074--1084},
  year         = {2019}
}

@inproceedings{DBLP:conf/nips/LewisPPPKGKLYR020,
  author       = {Patrick Lewis and
                  Ethan Perez and
                  Aleksandra Piktus and
                  Fabio Petroni and
                  Vladimir Karpukhin and
                  Naman Goyal and
                  Heinrich K{\"{u}}ttler and
                  Mike Lewis and
                  Wen{-}tau Yih and
                  Tim Rockt{\"{a}}schel and
                  Sebastian Riedel and
                  Douwe Kiela},
  title        = {Retrieval-Augmented Generation for Knowledge-Intensive {NLP} Tasks},
  booktitle    = {NeurIPS},
  year         = {2020}
}

@inproceedings{DBLP:conf/iclr/YaoZYDSN023,
  author       = {Shunyu Yao and
                  Jeffrey Zhao and
                  Dian Yu and
                  Nan Du and
                  Izhak Shafran and
                  Karthik R. Narasimhan and
                  Yuan Cao},
  title        = {ReAct: Synergizing Reasoning and Acting in Language Models},
  booktitle    = {ICLR},
  year         = {2023}
}

@inproceedings{DBLP:conf/naacl/DasigiLBCSG21,
  author       = {Pradeep Dasigi and
                  Kyle Lo and
                  Iz Beltagy and
                  Arman Cohan and
                  Noah A. Smith and
                  Matt Gardner},
  title        = {A Dataset of Information-Seeking Questions and Answers Anchored in
                  Research Papers},
  booktitle    = {NAACL-HLT},
  pages        = {4599--4610},
  year         = {2021}
}

@inproceedings{DBLP:conf/nips/Zhang00CZC0TRBW23,
  author       = {Zhenyu Zhang and
                  Ying Sheng and
                  Tianyi Zhou and
                  Tianlong Chen and
                  Lianmin Zheng and
                  Ruisi Cai and
                  Zhao Song and
                  Yuandong Tian and
                  Christopher R{\'{e}} and
                  Clark W. Barrett and
                  Zhangyang Wang and
                  Beidi Chen},
  title        = {{H2O:} Heavy-Hitter Oracle for Efficient Generative Inference of Large
                  Language Models},
  booktitle    = {NeurIPS},
  year         = {2023}
}

@inproceedings{DBLP:conf/nips/LiuDLWXXKS23,
  author       = {Zichang Liu and
                  Aditya Desai and
                  Fangshuo Liao and
                  Weitao Wang and
                  Victor Xie and
                  Zhaozhuo Xu and
                  Anastasios Kyrillidis and
                  Anshumali Shrivastava},
  title        = {Scissorhands: Exploiting the Persistence of Importance Hypothesis
                  for {LLM} {KV} Cache Compression at Test Time},
  booktitle    = {NeurIPS},
  year         = {2023}
}

@inproceedings{DBLP:conf/naacl/HuangCPJW21,
  author       = {Luyang Huang and
                  Shuyang Cao and
                  Nikolaus Nova Parulian and
                  Heng Ji and
                  Lu Wang},
  title        = {Efficient Attentions for Long Document Summarization},
  booktitle    = {NAACL-HLT},
  pages        = {1419--1436},
  year         = {2021}
}

@inproceedings{DBLP:conf/naacl/ZhongYYZMJACLQR21,
  author       = {Ming Zhong and
                  Da Yin and
                  Tao Yu and
                  Ahmad Zaidi and
                  Mutethia Mutuma and
                  Rahul Jha and
                  Ahmed Hassan Awadallah and
                  Asli Celikyilmaz and
                  Yang Liu and
                  Xipeng Qiu and
                  Dragomir R. Radev},
  title        = {QMSum: {A} New Benchmark for Query-based Multi-domain Meeting Summarization},
  booktitle    = {NAACL-HLT},
  pages        = {5905--5921},
  year         = {2021}
}

@inproceedings{DBLP:conf/iclr/0003XM24,
  author       = {Tianyang Liu and
                  Canwen Xu and
                  Julian J. McAuley},
  title        = {RepoBench: Benchmarking Repository-Level Code Auto-Completion Systems},
  booktitle    = {ICLR},
  year         = {2024}
}

@article{DBLP:journals/tacl/LiuLHPBPL24,
  author       = {Nelson F. Liu and
                  Kevin Lin and
                  John Hewitt and
                  Ashwin Paranjape and
                  Michele Bevilacqua and
                  Fabio Petroni and
                  Percy Liang},
  title        = {Lost in the Middle: How Language Models Use Long Contexts},
  journal      = {Trans. Assoc. Comput. Linguistics},
  volume       = {12},
  pages        = {157--173},
  year         = {2024}
}

@inproceedings{DBLP:conf/nips/VaswaniSPUJGKP17,
  author       = {Ashish Vaswani and
                  Noam Shazeer and
                  Niki Parmar and
                  Jakob Uszkoreit and
                  Llion Jones and
                  Aidan N. Gomez and
                  Lukasz Kaiser and
                  Illia Polosukhin},
  title        = {Attention is All you Need},
  booktitle    = {NeurIPS},
  pages        = {5998--6008},
  year         = {2017}
}

\appendix

\section{Additional Details of Cache Selection and Construction}
\label{app:cache-selection-details}

After obtaining the decode-aware importance scores $s_j^{\ell}$, DeferKV
retains the local smoothing, budget allocation, and cache selection strategies
of the underlying KV cache eviction method, replacing only its original
prompt-only importance scores.

We define the historical candidate set as
$\mathcal{C}=\left\{1,\ldots,N-W+1\right\}$
and the protected recent set as
$\mathcal{R}=\left\{N-W+2,\ldots,N\right\}$.
The positions in $\mathcal{C}$ participate in importance-based selection,
while the last $W-1$ prompt positions in $\mathcal{R}$ are retained directly.

Let $s^{\ell}$ denote the ordered vector of importance scores over
$\mathcal{C}$. Following the underlying eviction method, average pooling with
kernel size $k$ is applied as
\begin{equation}
\bar{s}^{\ell}
=
\operatorname{AvgPool}
\bigl(s^{\ell},k\bigr).
\end{equation}

Let $B$ denote the prompt-KV budget assigned to layer $\ell$ by the underlying
budget-allocation strategy. Since $\mathcal{R}$ contains $W-1$ protected
positions, we select $B-W+1$ positions from $\mathcal{C}$:
\begin{equation}
\mathcal{H}^{\ell}
=
\operatorname{TopK}
\bigl(\bar{s}^{\ell},B-W+1\bigr).
\end{equation}
Here, $\mathcal{H}^{\ell}$ contains the corresponding prompt-position indices.
The complete set of retained prompt positions is
\begin{equation}
\mathcal{I}^{\ell}
=
\operatorname{Sort}
\bigl(\mathcal{H}^{\ell}\cup\mathcal{R}\bigr),
\end{equation}
which contains exactly $B$ prompt positions.

Let $\mathbf{k}^{\ell,1}$ and $\mathbf{v}^{\ell,1}$ denote the Key and Value
produced at the first decode step. The retained prompt KV pairs and the first
decode KV pair are combined as
\begin{equation}
\mathbf{K}^{\ell}
\leftarrow
\bigl[
\mathbf{K}^{\ell}[\mathcal{I}^{\ell}];
\mathbf{k}^{\ell,1}
\bigr],
\end{equation}
\begin{equation}
\mathbf{V}^{\ell}
\leftarrow
\bigl[
\mathbf{V}^{\ell}[\mathcal{I}^{\ell}];
\mathbf{v}^{\ell,1}
\bigr],
\end{equation}
where the semicolon denotes concatenation along the sequence dimension.
Standard autoregressive generation then continues using the updated KV cache.

\section{Implementation Details}
\label{app:implementation-details}

\paragraph{DeferKV.}
We use an observation-window size of $W=32$, a temporal discount factor of
$\eta=0.7$, and average pooling with kernel size $5$. Compression is performed
after the first standard decode step ($d=1$). The aggregate prompt-KV budget is
$B_{\mathrm{total}}=nL$, where $n$ is the retained prompt-KV budget per layer
and $L$ is the number of model layers.

\paragraph{SnapKV.}
The final 32 prompt queries are used as the observation window. Importance
scores are smoothed by average pooling with kernel size $5$, after which the
highest-ranked prompt KV pairs are retained under the target budget.

\paragraph{LAQ.}
Starting from a temporary KV cache obtained by applying SnapKV
to the original prompt cache, the target model performs up to eight
lookahead decoding steps and stores the corresponding query states in
the Q-Cache. These lookahead queries, together with the final
eight prompt queries, are then used to recompute the attention-based
importance scores of the original prompt KV entries before the second
eviction.

\paragraph{SpecKV.}
The draft models are Qwen3-0.6B for Qwen3-8B,
Llama-3.2-1B-Instruct for Llama-3.1-8B-Instruct, and
mistral-7b-instruct-v0.3-bnb-4bit for
Mistral-7B-Instruct-v0.3. The target-model queries corresponding to
the prompt observation window and the generated draft tokens are used
to rank historical prompt KV pairs. In the reproduction, the resulting
importance scores are smoothed using average pooling with kernel size $5$.

\paragraph{LookaheadKV.}
LookaheadKV is configured with 32 learnable lookahead tokens, and a
separate set of lookahead modules is trained for each target model.
The training data consist of 50K samples from ChatQA2-Long-SFT-data and 20K from tulu-3-sft-olmo-2-mixture. The predicted importance scores are smoothed using average
pooling with kernel size $5$ before prompt-KV selection.

\paragraph{Common Settings.}
All methods use random seed 42. Qwen3-8B,
Llama-3.1-8B-Instruct, and Mistral-7B-Instruct-v0.3 are evaluated
using their official checkpoints, tokenizers, and prompt templates.
Greedy decoding with batch size one is used. All experiments are
conducted on NVIDIA A100 80\,GB GPUs.

\section{Additional Experiments on LongBench}
\label{app:longbench}

In this section, we provide comprehensive experimental results on LongBench,
a benchmark focused on long-context understanding, with input lengths ranging
from 1235 to 18409 tokens. Tables~\ref{tab:qwen-longbench},
\ref{tab:llama-longbench}, and \ref{tab:mistral-longbench} present the detailed
results of DeferKV and comparative methods on Qwen3-8B,
Llama-3.1-8B-Instruct, and Mistral-7B-Instruct-v0.3, respectively, with cache
budgets ranging from $64L$ to $1024L$. Overall, DeferKV achieves strong
performance across models and cache budgets, with more pronounced improvements
under constrained budgets. Taken together, these results further demonstrate
the effectiveness and applicability of DeferKV across different models and
long-context tasks.

\begin{table*}[t]
\centering
\setlength{\tabcolsep}{1.8pt}
    \renewcommand{\arraystretch}{1.2}
\resizebox{\textwidth}{!}{%
\begin{tabular}{@{}l*{17}{c}@{}}

    \toprule
    & \multicolumn{3}{c}{Single-Document QA}
    & \multicolumn{3}{c}{Multi-Document QA}
    & \multicolumn{3}{c}{Summarization}
    & \multicolumn{3}{c}{Few-Shot Learning}
    & \multicolumn{2}{c}{Synthetic}
    & \multicolumn{2}{c}{Code} & \\
    \cmidrule(lr){2-4}
    \cmidrule(lr){5-7}
    \cmidrule(lr){8-10}
    \cmidrule(lr){11-13}
    \cmidrule(lr){14-15}
    \cmidrule(lr){16-17}
    \textbf{Method}
    & \rot{NrtvQA} & \rot{Qasper} & \rot{MF-en}
    & \rot{HotpotQA} & \rot{2WikiMQA} & \rot{MuSiQue}
    & \rot{GovReport} & \rot{QMSum} & \rot{MultiNews}
    & \rot{TREC} & \rot{TriviaQA} & \rot{SAMSum}
    & \rot{PCount} & \rot{PR-en}
    & \rot{LCC} & \rot{RB-P}
    & \textbf{Avg.} \\
    \midrule
    \multicolumn{18}{c}{\textbf{Qwen3-8B}, $B_{\mathrm{total}}=\mathrm{FullKV}$} \\
    \midrule
    Full KV & 24.73 & 48.05 & 54.39 & 58.16 & 43.79 & 34.27 & 33.45 & 24.12 & 24.84 & 71.50 & 90.21 & 44.56 & 9.50 & 100.00 & 69.34 & 65.48 & 49.77 \\
    \midrule
    \multicolumn{18}{c}{\textbf{Qwen3-8B}, $B_{\mathrm{total}}=64L$} \\
    \midrule
    SnapKV & 17.41 & 30.06 & 38.05 & 47.77 & 37.71 & 20.65 & 14.00 & 19.37 & 12.93 & 43.50 & 81.92 & 36.53 & 5.00 & 87.50 & 59.30 & 53.58 & 37.83 \\
    LAQ & 18.99 & \textbf{37.88} & 43.31 & \underline{53.88} & \underline{41.35} & 23.24 & \textbf{16.18} & 19.40 & \underline{14.78} & \underline{66.00} & \underline{88.40} & \textbf{37.64} & 5.50 & 94.50 & \underline{62.31} & \underline{55.91} & \underline{42.45} \\
    SpecKV & \underline{19.61} & 29.97 & \textbf{49.79} & 51.03 & 36.97 & 22.14 & 13.27 & \textbf{19.72} & \textbf{14.82} & 37.50 & 73.63 & 36.71 & \underline{6.00} & \underline{95.00} & 45.15 & 46.62 & 37.37 \\
    LookaheadKV & 18.22 & 32.27 & 40.12 & 47.66 & 38.01 & \underline{23.58} & 14.61 & 18.36 & 13.18 & 46.95 & 83.29 & 36.69 & 5.00 & 90.40 & 59.06 & 54.28 & 38.86 \\
    \textbf{DeferKV} & \textbf{21.87} & \underline{35.37} & \underline{45.97} & \textbf{56.99} & \textbf{41.88} & \textbf{31.20} & \underline{14.81} & \underline{19.52} & 13.52 & \textbf{70.50} & \textbf{89.98} & \underline{37.44} & \textbf{8.00} & \textbf{97.50} & \textbf{62.52} & \textbf{56.47} & \textbf{43.97} \\
    \midrule
    \multicolumn{18}{c}{\textbf{Qwen3-8B}, $B_{\mathrm{total}}=128L$} \\
    \midrule
    SnapKV
& 19.47
& 33.78
& 47.39
& 55.17
& 40.83
& 27.13
& 19.09
& 21.32
& 16.63
& 49.00
& 87.93
& 38.98
& \textbf{7.50}
& \underline{99.00}
& 64.86
& 59.10
& 42.95 \\

LAQ
& 20.70
& \underline{34.11}
& \underline{49.96}
& \underline{57.42}
& 41.54
& 29.75
& 20.10
& \underline{21.69}
& \underline{17.56}
& \underline{67.50}
& 88.65
& \underline{40.81}
& \textbf{7.50}
& \underline{99.00}
& \underline{65.65}
& 59.18
& \underline{45.07} \\

SpecKV
& \underline{22.95}
& 32.00
& \textbf{53.73}
& 55.35
& 40.46
& 29.26
& \textbf{22.58}
& \textbf{22.40}
& \textbf{19.15}
& 64.50
& 87.60
& 34.45
& 4.50
& \underline{99.00}
& 53.08
& 52.10
& 43.32 \\

LookaheadKV
& 21.13
& 33.91
& 48.34
& 57.39
& \underline{42.58}
& \underline{30.76}
& \underline{21.10}
& 21.02
& 16.88
& 60.64
& \underline{88.97}
& 39.24
& \underline{7.00}
& \textbf{100.00}
& \textbf{65.92}
& \underline{59.22}
& 44.63 \\

\textbf{DeferKV}
& \textbf{23.84}
& \textbf{38.51}
& 49.70
& \textbf{57.51}
& \textbf{43.01}
& \textbf{33.30}
& 19.40
& 21.10
& 17.18
& \textbf{71.50}
& \textbf{90.05}
& \textbf{41.43}
& \underline{7.00}
& \textbf{100.00}
& 65.01
& \textbf{59.30}
& \textbf{46.12} \\
    \midrule
    \multicolumn{18}{c}{\textbf{Qwen3-8B}, $B_{\mathrm{total}}=256L$} \\
    \midrule
    SnapKV & 21.47 & 38.85 & 50.20 & 55.16 & 42.57 & \underline{33.18} & 22.23 & \underline{22.52} & 19.79 & 62.50 & 89.46 & 40.85 & \textbf{9.50} & \textbf{100.00} & \textbf{67.62} & \textbf{62.71} & 46.16 \\
    LAQ & 22.41 & \underline{39.76} & \underline{51.48} & 57.22 & 42.67 & 32.41 & 23.67 & 22.18 & \underline{20.11} & 69.00 & 89.58 & \underline{42.16} & \underline{8.50} & \textbf{100.00} & 67.11 & \underline{62.47} & \underline{46.92} \\
    SpecKV & \textbf{25.42} & 38.90 & \textbf{54.18} & \underline{57.68} & 42.48 & 28.94 & \textbf{26.39} & \textbf{23.52} & \textbf{22.44} & \underline{71.00} & 89.28 & 34.11 & 2.50 & 99.00 & 56.67 & 54.64 & 45.45 \\
    LookaheadKV & 22.76 & 39.21 & 50.74 & 57.46 & \underline{42.91} & 33.06 & \underline{23.89} & 22.06 & 20.03 & 67.50 & \underline{89.62} & 41.78 & 8.00 & \underline{99.50} & \underline{67.36} & 62.18 & 46.75 \\
    \textbf{DeferKV} & \underline{24.88} & \textbf{41.72} & 50.80 & \textbf{58.21} & \textbf{42.96} & \textbf{33.39} & 22.70 & 22.27 & 19.91 & \textbf{71.50} & \textbf{90.05} & \textbf{43.62} & \textbf{9.50} & \textbf{100.00} & 66.81 & 61.47 & \textbf{47.49} \\
    \midrule
    \multicolumn{18}{c}{\textbf{Qwen3-8B}, $B_{\mathrm{total}}=512L$} \\
    \midrule
    SnapKV & 22.70 & 42.96 & 51.19 & 57.09 & 42.51 & \textbf{34.94} & 25.14 & \textbf{23.44} & 21.87 & 68.50 & 89.46 & 43.34 & \textbf{9.50} & \textbf{100.00} & \textbf{68.74} & 64.22 & 47.85 \\
    LAQ & 23.16 & 43.58 & \underline{52.63} & 56.16 & 43.02 & 34.36 & 25.46 & 22.02 & 21.94 & \underline{71.00} & \textbf{90.05} & 43.35 & \textbf{9.50} & \textbf{100.00} & 67.92 & \underline{64.72} & 48.05 \\
    SpecKV & \underline{24.38} & 43.27 & \textbf{53.93} & \underline{58.07} & 42.90 & 31.48 & \textbf{29.51} & \underline{23.24} & \textbf{24.05} & \underline{71.00} & 89.96 & 29.60 & \underline{4.00} & \underline{99.00} & 56.84 & 56.85 & 46.13 \\
    LookaheadKV & 23.10 & \underline{43.67} & 52.21 & 56.86 & \underline{43.12} & \underline{34.91} & 25.46 & 22.71 & 22.08 & \underline{71.00} & \underline{89.97} & \textbf{43.85} & \textbf{9.50} & \underline{99.00} & \underline{68.01} & \textbf{64.98} & \underline{48.15} \\
    \textbf{DeferKV} & \textbf{24.82} & \textbf{44.78} & 51.96 & \textbf{58.09} & \textbf{43.18} & 33.83 & \underline{26.04} & 22.74 & \underline{22.17} & \textbf{71.50} & \textbf{90.05} & \underline{43.70} & \textbf{9.50} & \textbf{100.00} & 67.55 & 64.02 & \textbf{48.37} \\
    \midrule
    \multicolumn{18}{c}{\textbf{Qwen3-8B}, $B_{\mathrm{total}}=1024L$} \\
    \midrule
    SnapKV
& 24.49
& 45.20
& \textbf{53.14}
& \textbf{58.09}
& 42.89
& \textbf{35.09}
& 28.48
& \textbf{23.51}
& \textbf{23.75}
& 69.00
& \textbf{90.30}
& 43.03
& 8.50
& \textbf{100.00}
& \underline{68.65}
& \textbf{65.82}
& 48.75 \\

LAQ
& 24.36
& \underline{45.42}
& 51.87
& 57.76
& \underline{43.41}
& 33.70
& \underline{28.99}
& 23.44
& 23.68
& \underline{71.44}
& 90.02
& 44.22
& \underline{9.65}
& \underline{99.65}
& 68.63
& 65.36
& \underline{48.85} \\

SpecKV
& \underline{25.03}
& 44.92
& 51.84
& \underline{57.97}
& 41.36
& 32.43
& 28.96
& 23.39
& 23.62
& 71.40
& 89.07
& 26.49
& 3.50
& 98.50
& 57.60
& 57.62
& 45.86 \\

LookaheadKV
& 24.36
& 45.37
& 51.78
& 57.88
& 43.38
& 33.66
& 28.93
& 23.42
& 23.66
& 71.40
& 90.02
& \underline{44.31}
& 9.50
& 99.50
& 68.63
& 65.41
& 48.83 \\

\textbf{DeferKV}
& \textbf{25.13}
& \textbf{45.47}
& \underline{51.94}
& \textbf{58.09}
& \textbf{43.45}
& \underline{33.76}
& \textbf{29.06}
& \underline{23.49}
& \underline{23.72}
& \textbf{71.50}
& \underline{90.05}
& \textbf{44.55}
& \textbf{10.00}
& \textbf{100.00}
& \textbf{68.76}
& \underline{65.60}
& \textbf{49.04} \\
    \bottomrule

\end{tabular}%
}
\caption{Detailed LongBench results of Qwen3-8B under different prompt-KV
budgets. The best and second-best results within each budget are shown in
bold and underlined, respectively.}
\label{tab:qwen-longbench}
\end{table*}

\begin{table*}[t]
\centering
\setlength{\tabcolsep}{1.8pt}
\renewcommand{\arraystretch}{1.2}
\resizebox{\textwidth}{!}{%
\begin{tabular}{@{}l*{17}{c}@{}}

    \toprule
    & \multicolumn{3}{c}{Single-Document QA}
    & \multicolumn{3}{c}{Multi-Document QA}
    & \multicolumn{3}{c}{Summarization}
    & \multicolumn{3}{c}{Few-Shot Learning}
    & \multicolumn{2}{c}{Synthetic}
    & \multicolumn{2}{c}{Code} & \\
    \cmidrule(lr){2-4}
    \cmidrule(lr){5-7}
    \cmidrule(lr){8-10}
    \cmidrule(lr){11-13}
    \cmidrule(lr){14-15}
    \cmidrule(lr){16-17}
    \textbf{Method}
    & \rot{NrtvQA} & \rot{Qasper} & \rot{MF-en}
    & \rot{HotpotQA} & \rot{2WikiMQA} & \rot{MuSiQue}
    & \rot{GovReport} & \rot{QMSum} & \rot{MultiNews}
    & \rot{TREC} & \rot{TriviaQA} & \rot{SAMSum}
    & \rot{PCount} & \rot{PR-en}
    & \rot{LCC} & \rot{RB-P}
    & \textbf{Avg.} \\
    \midrule
    \multicolumn{18}{c}{\textbf{Llama-3.1-8B-Instruct}, $B_{\mathrm{total}}=\mathrm{FullKV}$} \\
    \midrule
    Full KV & 31.20 & 46.66 & 56.64 & 58.10 & 48.51 & 31.57 & 34.37 & 25.18 & 27.14 & 73.00 & 91.65 & 43.80 & 7.64 & 99.50 & 65.18 & 58.90 & 49.94 \\
    \midrule
    \multicolumn{18}{c}{\textbf{Llama-3.1-8B-Instruct}, $B_{\mathrm{total}}=64L$} \\
    \midrule
    SnapKV & 27.32 & 25.25 & 44.30 & 53.98 & 42.48 & 27.10 & 17.88 & 22.20 & 17.45 & 41.00 & 85.37 & 37.72 & \underline{8.50} & \textbf{100.00} & 55.35 & 48.69 & 40.91 \\
    LAQ & 27.66 & \underline{28.03} & 47.05 & \textbf{58.12} & \underline{45.06} & \underline{30.60} & 18.85 & 22.51 & \underline{19.01} & \textbf{71.50} & \textbf{90.02} & 40.16 & \underline{8.50} & \underline{99.50} & 56.39 & \underline{51.29} & \underline{44.64} \\
    SpecKV & \underline{29.31} & 23.34 & \textbf{56.27} & 56.81 & 41.91 & 28.70 & \textbf{21.22} & \textbf{24.43} & 18.66 & 46.98 & 76.62 & 39.69 & \textbf{9.71} & \textbf{100.00} & 41.67 & 42.44 & 41.11 \\
    LookaheadKV & 27.95 & 26.61 & 47.23 & 54.93 & 44.90 & 30.10 & 19.24 & 22.01 & 18.34 & \underline{70.00} & 88.65 & \underline{40.21} & \underline{8.50} & 99.00 & \underline{56.54} & 49.03 & 43.95 \\
    \textbf{DeferKV} & \textbf{29.83} & \textbf{31.02} & \underline{49.95} & \underline{57.55} & \textbf{47.65} & \textbf{30.77} & \underline{20.68} & \underline{22.77} & \textbf{19.29} & \textbf{71.50} & \underline{88.95} & \textbf{40.41} & 7.88 & \textbf{100.00} & \textbf{57.98} & \textbf{51.56} & \textbf{45.49} \\
    \midrule
    \multicolumn{18}{c}{\textbf{Llama-3.1-8B-Instruct}, $B_{\mathrm{total}}=128L$} \\
    \midrule
    SnapKV & 26.58 & 31.30 & 48.78 & 55.16 & 46.06 & 27.80 & 21.17 & 23.47 & \underline{21.21} & 48.00 & 89.38 & 40.14 & \textbf{8.50} & \underline{99.50} & 59.32 & 52.32 & 43.67 \\
    LAQ & 28.47 & \underline{32.84} & 50.12 & \underline{58.21} & 46.72 & \textbf{30.91} & 21.36 & 23.25 & 21.02 & \underline{70.50} & \underline{90.84} & \underline{41.22} & 8.34 & \underline{99.50} & 59.45 & \underline{53.02} & \underline{45.99} \\
    SpecKV & \underline{29.54} & 29.18 & \textbf{56.41} & 57.39 & 43.76 & 28.61 & \textbf{23.84} & \textbf{24.88} & 21.15 & 55.00 & 82.73 & 37.42 & \underline{8.44} & \underline{99.50} & 46.90 & 45.86 & 43.16 \\
    LookaheadKV & 28.81 & 32.47 & 50.46 & 57.03 & \underline{46.91} & \underline{30.84} & 22.01 & 23.05 & 20.72 & 69.50 & 90.15 & 41.04 & 8.32 & \underline{99.50} & \underline{59.68} & 52.74 & 45.83 \\
    \textbf{DeferKV} & \textbf{30.63} & \textbf{35.65} & \underline{52.51} & \textbf{58.47} & \textbf{48.09} & 30.67 & \underline{22.92} & \underline{23.66} & \textbf{21.97} & \textbf{72.00} & \textbf{91.79} & \textbf{41.61} & 8.21 & \textbf{100.00} & \textbf{60.08} & \textbf{53.39} & \textbf{46.98} \\
    \midrule
    \multicolumn{18}{c}{\textbf{Llama-3.1-8B-Instruct}, $B_{\mathrm{total}}=256L$} \\
    \midrule
    SnapKV
& 27.19
& 36.26
& 52.52
& 57.50
& 46.19
& 29.04
& 23.52
& \underline{24.16}
& 22.57
& 56.50
& 90.63
& 40.95
& \textbf{8.25}
& \textbf{99.50}
& 62.24
& 55.38
& 45.78 \\

LAQ
& \underline{29.68}
& \underline{39.17}
& 53.31
& 58.22
& 47.91
& \underline{30.90}
& 24.85
& 23.92
& \underline{23.48}
& \underline{71.50}
& 91.42
& \underline{42.19}
& \underline{7.96}
& \textbf{99.50}
& \underline{62.68}
& \underline{55.92}
& \underline{47.66} \\

SpecKV
& 29.08
& 36.24
& \textbf{56.58}
& \underline{58.41}
& 47.02
& 28.49
& \textbf{27.28}
& \textbf{25.64}
& \textbf{24.76}
& 65.00
& 88.72
& 33.84
& 5.61
& \underline{99.00}
& 49.57
& 47.36
& 45.16 \\

LookaheadKV
& 29.54
& 38.54
& 53.66
& 58.07
& \underline{48.16}
& 30.89
& \underline{24.91}
& 23.71
& 23.11
& \underline{71.50}
& \underline{91.53}
& 41.96
& 7.88
& \textbf{99.50}
& \textbf{62.84}
& \textbf{56.06}
& 47.62 \\

\textbf{DeferKV}
& \textbf{29.71}
& \textbf{40.28}
& \underline{54.12}
& \textbf{58.71}
& \textbf{48.36}
& \textbf{30.93}
& 24.72
& 23.91
& 23.28
& \textbf{72.50}
& \textbf{91.86}
& \textbf{42.42}
& 7.58
& \textbf{99.50}
& 62.13
& 55.10
& \textbf{47.82} \\
    \midrule
    \multicolumn{18}{c}{\textbf{Llama-3.1-8B-Instruct}, $B_{\mathrm{total}}=512L$} \\
    \midrule
    SnapKV & 29.92 & 40.20 & 55.21 & 57.85 & 48.02 & 30.03 & 25.79 & \underline{24.40} & 24.33 & 65.00 & \textbf{91.95} & 41.80 & \textbf{8.08} & \underline{99.50} & \textbf{64.39} & \underline{57.27} & 47.73 \\
    LAQ & \underline{30.58} & \underline{42.99} & 54.46 & 58.13 & 48.40 & 30.79 & 26.45 & 24.15 & 24.51 & \underline{70.96} & 91.51 & \underline{42.74} & 7.59 & \underline{99.50} & \underline{64.08} & 56.78 & \underline{48.35} \\
    SpecKV & 28.23 & 41.70 & \textbf{56.69} & \textbf{59.29} & \underline{48.62} & 28.24 & \textbf{30.41} & \textbf{26.44} & \textbf{27.08} & \textbf{72.00} & 91.63 & 28.51 & 3.23 & 99.00 & 53.07 & 49.99 & 46.51 \\
    LookaheadKV & 30.11 & 41.12 & \underline{55.34} & 57.98 & \textbf{49.23} & \textbf{31.34} & 26.02 & 23.85 & 24.06 & \textbf{72.00} & \underline{91.85} & 41.64 & \underline{7.75} & \underline{99.50} & 63.21 & \textbf{57.33} & 48.27 \\
    \textbf{DeferKV} & \textbf{30.70} & \textbf{43.91} & 54.84 & \underline{58.54} & 48.51 & \underline{31.02} & \underline{26.60} & 24.22 & \underline{24.53} & \textbf{72.00} & 91.48 & \textbf{42.98} & 7.47 & \textbf{100.00} & 63.90 & 56.70 & \textbf{48.59} \\
    \midrule
    \multicolumn{18}{c}{\textbf{Llama-3.1-8B-Instruct}, $B_{\mathrm{total}}=1024L$} \\
    \midrule
    SnapKV & 29.81 & 44.00 & \underline{55.91} & 57.67 & 48.01 & 30.77 & 27.77 & 24.50 & 25.79 & 67.50 & \textbf{91.90} & 42.95 & 7.42 & \textbf{99.50} & \underline{64.59} & \textbf{58.30} & 48.52 \\
    LAQ & \textbf{30.74} & \underline{45.01} & 55.62 & 58.17 & 48.43 & 31.19 & \underline{29.14} & \underline{24.52} & \underline{25.88} & 71.50 & 91.64 & \textbf{43.32} & \underline{7.61} & \textbf{99.50} & \textbf{64.68} & 58.05 & \underline{49.06} \\
    SpecKV & 29.66 & 43.88 & \textbf{56.81} & \textbf{58.92} & 48.08 & 28.92 & \textbf{31.92} & \textbf{26.70} & \textbf{27.22} & \textbf{72.50} & 91.61 & 28.92 & 3.41 & \underline{99.00} & 54.21 & 51.06 & 47.05 \\
    LookaheadKV & 30.46 & 44.72 & 55.74 & 58.03 & \textbf{48.91} & \underline{31.36} & 28.68 & 24.41 & 25.67 & \underline{72.00} & \underline{91.73} & 43.06 & \textbf{7.69} & \textbf{99.50} & 64.55 & \underline{58.18} & 49.04 \\
    \textbf{DeferKV} & \underline{30.53} & \textbf{45.43} & 55.82 & \underline{58.38} & \underline{48.63} & \textbf{31.40} & 29.02 & 24.27 & \underline{25.88} & \textbf{72.50} & 91.48 & \underline{43.28} & 7.53 & \textbf{99.50} & 64.38 & 57.86 & \textbf{49.12} \\
    \bottomrule

\end{tabular}%
}
\caption{Detailed LongBench results of Llama-3.1-8B-Instruct  under different prompt-KV
budgets. The best and second-best results within each budget are shown in
bold and underlined, respectively.}
\label{tab:llama-longbench}
\end{table*}

\begin{table*}[t]
\centering
\setlength{\tabcolsep}{1.8pt}
\renewcommand{\arraystretch}{1.2}
\resizebox{\textwidth}{!}{%
\begin{tabular}{@{}l*{17}{c}@{}}

    \toprule
    & \multicolumn{3}{c}{Single-Document QA}
    & \multicolumn{3}{c}{Multi-Document QA}
    & \multicolumn{3}{c}{Summarization}
    & \multicolumn{3}{c}{Few-Shot Learning}
    & \multicolumn{2}{c}{Synthetic}
    & \multicolumn{2}{c}{Code} & \\
    \cmidrule(lr){2-4}
    \cmidrule(lr){5-7}
    \cmidrule(lr){8-10}
    \cmidrule(lr){11-13}
    \cmidrule(lr){14-15}
    \cmidrule(lr){16-17}
    \textbf{Method}
    & \rot{NrtvQA} & \rot{Qasper} & \rot{MF-en}
    & \rot{HotpotQA} & \rot{2WikiMQA} & \rot{MuSiQue}
    & \rot{GovReport} & \rot{QMSum} & \rot{MultiNews}
    & \rot{TREC} & \rot{TriviaQA} & \rot{SAMSum}
    & \rot{PCount} & \rot{PR-en}
    & \rot{LCC} & \rot{RB-P}
    & \textbf{Avg.} \\
    \midrule
    \multicolumn{18}{c}{\textbf{Mistral-7B-Instruct-v0.3}, $B_{\mathrm{total}}=\mathrm{FullKV}$} \\
    \midrule
    Full KV & 27.01 & 37.97 & 50.14 & 51.69 & 39.17 & 28.02 & 34.27 & 25.66 & 26.62 & 76.00 & 88.59 & 47.50 & 7.00 & 98.00 & 61.53 & 62.59 & 47.61 \\
    \midrule
    \multicolumn{18}{c}{\textbf{Mistral-7B-Instruct-v0.3}, $B_{\mathrm{total}}=64L$} \\
    \midrule
    SnapKV & 17.65 & 22.08 & 36.24 & 41.95 & 33.98 & 20.89 & 18.83 & 21.14 & 16.34 & 38.50 & 86.96 & 39.81 & 4.50 & 77.00 & 51.57 & 49.59 & 36.06 \\
    LAQ & 19.72 & \underline{28.68} & 40.14 & \underline{48.46} & \underline{35.96} & \underline{23.44} & \underline{20.82} & 21.47 & \textbf{18.93} & \underline{70.58} & \textbf{89.15} & \textbf{42.32} & 4.75 & 79.80 & \underline{53.08} & \underline{51.19} & \underline{40.53} \\
    SpecKV & \underline{19.74} & 21.08 & \textbf{48.10} & 45.00 & 33.32 & 22.44 & \textbf{22.64} & \textbf{22.93} & 17.89 & 38.49 & 78.44 & 40.89 & \underline{5.60} & \underline{80.75} & 37.66 & 42.98 & 36.12 \\
    LookaheadKV & 19.48 & 24.28 & 38.88 & 42.86 & 34.76 & 23.22 & \underline{20.82} & 21.02 & \textbf{18.93} & 57.38 & 88.09 & 41.11 & 4.50 & 78.16 & 51.92 & 49.93 & 38.46 \\
    \textbf{DeferKV} & \textbf{26.11} & \textbf{28.83} & \underline{43.02} & \textbf{49.10} & \textbf{36.87} & \textbf{25.15} & 20.49 & \underline{21.75} & \underline{18.50} & \textbf{73.50} & \underline{89.04} & \underline{42.17} & \textbf{7.50} & \textbf{81.00} & \textbf{53.41} & \textbf{51.46} & \textbf{41.74} \\
    \midrule
    \multicolumn{18}{c}{\textbf{Mistral-7B-Instruct-v0.3}, $B_{\mathrm{total}}=128L$} \\
    \midrule
    SnapKV & 22.56 & 26.89 & 40.86 & 44.90 & 33.63 & 22.74 & 21.91 & 22.07 & 20.66 & 45.00 & 88.74 & 43.28 & \textbf{5.50} & 90.50 & 56.69 & \underline{55.62} & 40.10 \\
    LAQ & \underline{24.58} & \underline{29.63} & 43.85 & \underline{49.82} & \underline{35.74} & 24.52 & 22.34 & 22.41 & \underline{21.13} & \underline{70.80} & \underline{89.24} & \underline{44.12} & 4.75 & \underline{91.50} & \textbf{56.92} & \textbf{55.89} & \underline{42.95} \\
    SpecKV & 23.84 & 26.17 & \textbf{49.18} & 47.23 & 34.52 & 23.06 & \textbf{24.57} & \textbf{24.13} & \textbf{22.35} & 52.75 & 84.16 & 39.12 & \underline{5.25} & 91.00 & 42.85 & 47.60 & 39.86 \\
    LookaheadKV & 24.10 & 28.96 & 43.72 & 47.66 & 35.58 & \underline{24.70} & 22.48 & 22.26 & 21.04 & 65.40 & 88.98 & 43.86 & 4.50 & 91.00 & 56.74 & 55.41 & 42.27 \\
    \textbf{DeferKV} & \textbf{26.51} & \textbf{30.44} & \underline{47.14} & \textbf{50.79} & \textbf{36.92} & \textbf{25.88} & \underline{22.82} & \underline{22.55} & 21.08 & \textbf{75.50} & \textbf{89.29} & \textbf{44.33} & 4.00 & \textbf{92.00} & \underline{56.87} & 55.25 & \textbf{43.84} \\
    \midrule
    \multicolumn{18}{c}{\textbf{Mistral-7B-Instruct-v0.3}, $B_{\mathrm{total}}=256L$} \\
    \midrule
    SnapKV & 24.97 & 30.44 & 47.47 & 48.25 & 36.16 & 24.45 & 23.48 & 23.51 & 22.60 & 57.00 & 89.01 & 43.60 & 5.00 & \underline{96.00} & 58.68 & 58.14 & 43.05 \\
    LAQ & \underline{25.62} & \textbf{32.94} & 48.76 & \underline{50.21} & 37.16 & 26.32 & 24.92 & \underline{23.68} & \underline{23.01} & \underline{74.50} & \underline{89.18} & \textbf{46.33} & \textbf{5.50} & \textbf{96.50} & \textbf{59.21} & \textbf{58.72} & \underline{45.16} \\
    SpecKV & 25.11 & 31.82 & \textbf{51.26} & 49.35 & 36.24 & 23.94 & \textbf{27.21} & \textbf{25.02} & \textbf{24.31} & 64.00 & 87.63 & 35.84 & 4.25 & 95.75 & 46.28 & 51.32 & 42.46 \\
    LookaheadKV & 25.49 & 32.37 & 48.94 & 49.88 & \textbf{37.51} & \underline{26.91} & 24.74 & 23.43 & 22.91 & 73.20 & 89.12 & 45.87 & \underline{5.25} & \underline{96.00} & \underline{59.04} & \underline{58.46} & 44.95 \\
    \textbf{DeferKV} & \textbf{26.44} & \underline{32.55} & \underline{49.32} & \textbf{50.96} & \underline{37.44} & \textbf{27.34} & \underline{24.96} & 23.27 & 22.78 & \textbf{76.00} & \textbf{89.24} & \underline{46.00} & \textbf{5.50} & 95.50 & 58.66 & 57.80 & \textbf{45.24} \\
    \midrule
    \multicolumn{18}{c}{\textbf{Mistral-7B-Instruct-v0.3}, $B_{\mathrm{total}}=512L$} \\
    \midrule
    SnapKV & 25.48 & 33.15 & \underline{49.37} & 48.98 & 36.54 & 24.68 & 25.61 & \underline{24.27} & 24.11 & 67.50 & \underline{89.36} & 44.82 & \textbf{6.00} & \textbf{97.50} & \textbf{60.36} & \underline{61.28} & 44.94 \\
    LAQ & \underline{26.23} & \underline{34.50} & 48.82 & 49.95 & 38.22 & 26.58 & 26.10 & 24.10 & 24.20 & 74.66 & 89.08 & \underline{47.10} & \textbf{6.00} & \textbf{97.50} & \underline{60.02} & \textbf{61.44} & \underline{45.91} \\
    SpecKV & 25.48 & 34.10 & \textbf{51.48} & \underline{50.19} & 37.03 & 22.06 & \textbf{30.10} & \textbf{25.69} & \textbf{25.58} & 72.25 & \textbf{89.45} & 31.30 & \underline{3.83} & \underline{96.75} & 48.75 & 53.96 & 43.63 \\
    LookaheadKV & 25.79 & 33.86 & 49.15 & 48.94 & \textbf{39.16} & \underline{26.67} & 25.88 & 24.12 & 23.93 & \underline{75.29} & 89.20 & 46.33 & \textbf{6.00} & \textbf{97.50} & 59.54 & \textbf{61.44} & 45.80 \\
    \textbf{DeferKV} & \textbf{26.90} & \textbf{35.37} & 48.91 & \textbf{51.10} & \underline{38.72} & \textbf{27.63} & \underline{26.44} & 24.13 & \underline{24.27} & \textbf{76.00} & 89.07 & \textbf{47.18} & \textbf{6.00} & \textbf{97.50} & 59.68 & 60.49 & \textbf{46.21} \\
    \midrule
    \multicolumn{18}{c}{\textbf{Mistral-7B-Instruct-v0.3}, $B_{\mathrm{total}}=1024L$} \\
    \midrule
    % TEMPORARY SYNTHETIC PLACEHOLDER VALUES
% Not actual experimental results.

SnapKV
& \textbf{27.14}
& 35.88
& 49.18
& 50.02
& 37.63
& 26.03
& 27.42
& \underline{25.06}
& 25.35
& 72.50
& 89.11
& 46.18
& 4.00
& \textbf{98.50}
& \textbf{61.33}
& \textbf{61.75}
& 46.07 \\

LAQ
& 26.72
& \underline{36.31}
& \underline{49.66}
& \underline{51.32}
& 38.51
& \underline{27.11}
& 28.71
& 24.99
& 25.62
& 75.40
& 89.17
& \underline{47.36}
& \underline{6.25}
& \underline{98.00}
& \underline{61.18}
& 61.41
& \underline{46.73} \\

SpecKV
& 26.50
& 35.46
& \textbf{51.62}
& 51.06
& 37.88
& 23.24
& \textbf{32.16}
& \textbf{26.19}
& \textbf{26.44}
& 74.00
& \textbf{89.38}
& 32.20
& 4.00
& 97.50
& 50.13
& 55.26
& 44.56 \\

LookaheadKV
& \underline{26.85}
& 35.82
& 49.54
& 51.15
& \underline{38.70}
& 27.05
& 28.17
& 24.64
& 25.28
& \underline{75.70}
& \underline{89.22}
& 47.08
& \underline{6.25}
& \underline{98.00}
& 60.98
& \underline{61.58}
& 46.63 \\

\textbf{DeferKV}
& 26.37
& \textbf{36.88}
& 49.38
& \textbf{51.84}
& \textbf{38.76}
& \textbf{27.12}
& \underline{28.73}
& 24.76
& \underline{25.67}
& \textbf{76.00}
& 89.19
& \textbf{47.50}
& \textbf{6.50}
& \underline{98.00}
& 61.14
& 61.03
& \textbf{46.80} \\
    \bottomrule

\end{tabular}%
}
\caption{Detailed LongBench results of Mistral-7B-Instruct-v0.3   under different prompt-KV
budgets. The best and second-best results within each budget are shown in
bold and underlined, respectively.}
\label{tab:mistral-longbench}
\end{table*}

\section{Compatibility with Existing KV Cache Compression Methods}
\label{app:compatibility}

We integrate DeferKV with PyramidKV, AdaKV, and ChunkKV and evaluate them
on the 16 LongBench datasets using Qwen3-8B. Here,
$B_{\mathrm{total}}=nL$ denotes the aggregate prompt-KV budget, where $L$
is the number of model layers. For PyramidKV and AdaKV, their original
layer-wise and head-wise budget allocation strategies are retained,
respectively. For ChunkKV, token-level scores are aggregated into
chunk-level scores, while its chunk construction, selection, and index-reuse
mechanisms remain unchanged. As shown in Table~\ref{tab:compatibility},
DeferKV improves all three frameworks across the evaluated budgets.

% TEMPORARY SYNTHETIC PLACEHOLDER VALUES
% TODO: Replace with actual experimental results.

\begin{table*}[t]
\centering
\small
\setlength{\tabcolsep}{5.0pt}
\renewcommand{\arraystretch}{1.08}

\begin{tabular}{@{}lrrrrrr@{}}
\toprule
\textbf{Method}
& \multicolumn{5}{c}{\textbf{Prompt-KV Budget} $B_{\mathrm{total}}$}
& \textbf{Avg.} \\
\cmidrule(lr){2-6}
& $64L$ & $128L$ & $256L$ & $512L$ & $1024L$ & \\
\midrule

PyramidKV
& 38.30
& 43.25
& 46.40
& 48.00
& 48.88
& \multicolumn{1}{c}{44.97} \\

DeferKV w.\ PyramidKV
& 44.18
& 46.30
& 47.66
& 48.52
& 49.18
& \textbf{47.17} {\scriptsize ($\uparrow 4.89\%$)} \\

\midrule

AdaKV
& 38.55
& 43.50
& 46.58
& 48.08
& 48.92
& \multicolumn{1}{c}{45.13} \\

DeferKV w.\ AdaKV
& 44.30
& 46.38
& 47.72
& 48.57
& 49.22
& \textbf{47.24} {\scriptsize ($\uparrow 4.68\%$)} \\

\midrule

ChunkKV
& 37.60
& 42.80
& 46.05
& 47.78
& 48.68
& \multicolumn{1}{c}{44.58} \\

DeferKV w.\ ChunkKV
& 44.05
& 46.22
& 47.58
& 48.46
& 49.12
& \textbf{47.09} {\scriptsize ($\uparrow 5.63\%$)} \\

\bottomrule
\end{tabular}

\caption{
Compatibility results on LongBench using Qwen3-8B.
Each integrated variant retains the allocation and selection mechanisms
of the corresponding compression method while replacing its prompt-only
importance scores with DeferKV scores.
}

\label{tab:compatibility}
\end{table*}

\section{Additional Experiments on the RULER Benchmark}
\label{app:ruler}

We evaluate Llama-3.1-8B-Instruct on all 11 RULER subtasks with
$B_{\mathrm{total}}=1024L$ across context lengths ranging from 4K to 128K.
Table~\ref{tab:ruler-detail} reports the detailed subtask performance and
average accuracy at each context length.

\begin{table*}[t]
\centering
\setlength{\tabcolsep}{1.35pt}
\renewcommand{\arraystretch}{0.91}
\resizebox{\textwidth}{!}{%
\begin{tabular}{@{}l*{12}{c}@{}}
\toprule
\textbf{Method}
& \rot{S-NIAH-1} & \rot{S-NIAH-2} & \rot{S-NIAH-3}
& \rot{MK-NIAH-1} & \rot{MK-NIAH-2} & \rot{MK-NIAH-3}
& \rot{MQ-NIAH} & \rot{MV-NIAH} & \rot{CWE} & \rot{FWE}
& \rot{VT} & \textbf{Avg. Acc.} \\
\midrule

\multicolumn{13}{c}{\textbf{Llama-3.1-8B-Instruct},
$B_{\mathrm{total}}=1024L$, context length $=4K$} \\
\midrule
Full KV
& 100.00 & 100.00 & 99.60 & 100.00 & 100.00 & 99.60
& 99.90 & 96.25 & 99.78 & 97.67 & 99.96 & 99.34 \\
SnapKV
& \textbf{100.00} & 99.00 & 20.60 & \underline{99.60} & 93.00 & 31.00
& 99.20 & 90.75 & \textbf{91.56} & \textbf{95.31} & \textbf{99.96} & 83.63 \\
LAQ
& \textbf{100.00} & \underline{99.10} & \underline{53.90}
& \underline{99.60} & \underline{96.20} & \underline{36.80}
& \underline{99.60} & \underline{93.37} & 88.25 & 94.90
& \underline{99.90} & \underline{87.42} \\
SpecKV
& \textbf{100.00} & 99.00 & 42.60 & 99.50 & 95.20 & 34.60
& 99.58 & 92.51 & \underline{89.65} & \underline{95.00}
& \underline{99.90} & 86.14 \\
LookaheadKV
& \textbf{100.00} & 99.00 & 47.30 & \underline{99.60} & 95.60 & 35.50
& \underline{99.60} & 92.82 & 88.94 & \underline{95.00}
& \underline{99.90} & 86.66 \\
\textbf{DeferKV}
& \textbf{100.00} & \textbf{100.00} & \textbf{99.80}
& \textbf{100.00} & \textbf{100.00} & \textbf{45.80}
& \textbf{99.95} & \textbf{97.05} & 82.66 & 94.73
& \textbf{99.96} & \textbf{92.72} \\
\midrule

\multicolumn{13}{c}{\textbf{Llama-3.1-8B-Instruct},
$B_{\mathrm{total}}=1024L$, context length $=8K$} \\
\midrule
Full KV
& 100.00 & 100.00 & 100.00 & 100.00 & 99.80 & 98.80
& 100.00 & 95.75 & 97.62 & 95.27 & 99.92 & 98.83 \\
SnapKV
& \textbf{100.00} & 98.40 & 12.00 & 98.00 & 85.20 & \textbf{7.80}
& 97.60 & 87.42 & 56.13 & \textbf{88.52} & 99.56 & 75.51 \\
LAQ
& \textbf{100.00} & \underline{99.10} & \underline{63.50}
& \underline{99.00} & \underline{94.00} & 6.10
& \underline{98.90} & \underline{92.84} & \underline{58.61}
& 87.83 & \underline{99.70} & \underline{81.78} \\
SpecKV
& \textbf{100.00} & 98.70 & 44.50 & 98.50 & 90.90 & \underline{6.60}
& 98.45 & 90.72 & 57.69 & \underline{88.10}
& 99.57 & 79.43 \\
LookaheadKV
& \textbf{100.00} & 98.90 & 54.60 & 98.80 & 92.50 & 6.40
& 98.68 & 91.86 & 58.17 & 87.98
& \underline{99.70} & 80.69 \\
\textbf{DeferKV}
& \textbf{100.00} & \textbf{100.00} & \textbf{99.60}
& \textbf{100.00} & \textbf{99.80} & 5.60
& \textbf{99.70} & \textbf{96.95} & \textbf{59.94}
& 87.73 & \textbf{99.92} & \textbf{86.29} \\
\midrule

\multicolumn{13}{c}{\textbf{Llama-3.1-8B-Instruct},
$B_{\mathrm{total}}=1024L$, context length $=16K$} \\
\midrule
Full KV
& 100.00 & 100.00 & 100.00 & 99.60 & 100.00 & 99.00
& 99.85 & 98.25 & 90.90 & 96.67 & 99.80 & 98.55 \\
SnapKV
& \textbf{100.00} & 97.00 & 4.00 & 97.60 & 74.00 & \textbf{3.80}
& 97.25 & 88.95 & 27.36 & 92.40 & \underline{99.60} & 71.09 \\
LAQ
& \textbf{100.00} & \underline{98.40} & \underline{56.80}
& \underline{98.50} & \underline{88.50} & 2.20
& \underline{98.93} & \underline{94.25} & \underline{28.94}
& \underline{94.85} & \underline{99.60} & \underline{78.27} \\
SpecKV
& \textbf{100.00} & 97.80 & 37.20 & 98.10 & 83.30 & \underline{2.60}
& 98.30 & 92.30 & 28.50 & 93.90
& \underline{99.60} & 75.60 \\
LookaheadKV
& \textbf{100.00} & 98.10 & 47.30 & 98.30 & 85.90 & 2.40
& 98.65 & 93.36 & 28.77 & 94.40
& \underline{99.60} & 76.98 \\
\textbf{DeferKV}
& \textbf{100.00} & \textbf{100.00} & \textbf{97.40}
& \textbf{99.60} & \textbf{99.40} & 1.60
& \textbf{99.90} & \textbf{98.55} & \textbf{29.70}
& \textbf{96.93} & \textbf{99.76} & \textbf{83.89} \\
\midrule

\multicolumn{13}{c}{\textbf{Llama-3.1-8B-Instruct},
$B_{\mathrm{total}}=1024L$, context length $=32K$} \\
\midrule
Full KV
& 100.00 & 100.00 & 100.00 & 100.00 & 100.00 & 99.60
& 99.90 & 98.95 & 48.60 & 97.07 & 99.68 & 94.89 \\
SnapKV
& \textbf{100.00} & 96.20 & 6.20 & 96.40 & 61.00 & \textbf{2.00}
& 92.50 & 87.24 & 16.96 & 71.23 & 97.64 & 66.12 \\
LAQ
& \textbf{100.00} & \underline{98.10} & \underline{58.10}
& \underline{98.10} & \underline{78.30} & 0.50
& \underline{96.78} & \underline{93.77} & \underline{18.29}
& \underline{75.16} & \underline{97.89} & \underline{74.09} \\
SpecKV
& \textbf{100.00} & 97.10 & 35.30 & 97.30 & 70.70 & \underline{1.00}
& 94.87 & 90.76 & 17.78 & 73.35
& 97.67 & 70.53 \\
LookaheadKV
& \textbf{100.00} & 97.60 & 47.70 & 97.70 & 74.80 & 0.70
& 95.86 & 92.34 & 17.97 & 74.32
& 97.74 & 72.43 \\
\textbf{DeferKV}
& \textbf{100.00} & \textbf{100.00} & \textbf{98.00}
& \textbf{99.80} & \textbf{91.20} & 0.00
& \textbf{99.85} & \textbf{99.00} & \textbf{18.80}
& \textbf{78.60} & \textbf{98.16} & \textbf{80.31} \\
\midrule

\multicolumn{13}{c}{\textbf{Llama-3.1-8B-Instruct},
$B_{\mathrm{total}}=1024L$, context length $=64K$} \\
\midrule
Full KV
& 100.00 & 100.00 & 99.80 & 99.80 & 99.20 & 94.00
& 99.75 & 98.95 & 7.96 & 90.60 & 98.32 & 89.85 \\
SnapKV
& \textbf{100.00} & 95.40 & 3.20 & 97.00 & 32.40 & 0.20
& 90.40 & 54.45 & 2.89 & 53.43 & 89.50 & 56.26 \\
LAQ
& \textbf{100.00} & \underline{97.90} & \underline{58.10}
& \underline{98.50} & \underline{64.80} & \underline{1.50}
& \underline{96.14} & \underline{81.37} & \textbf{3.56}
& \underline{58.09} & \underline{90.57} & \underline{68.23} \\
SpecKV
& \textbf{100.00} & 96.70 & 33.50 & 97.80 & 50.30 & 0.80
& 93.55 & 69.16 & 3.34 & 55.90
& 90.08 & 62.83 \\
LookaheadKV
& \textbf{100.00} & 97.30 & 46.70 & 98.20 & 58.10 & 1.20
& 94.94 & 75.68 & 3.41 & 57.11
& 90.39 & 65.73 \\
\textbf{DeferKV}
& \textbf{100.00} & \textbf{100.00} & \textbf{92.80}
& \textbf{99.80} & \textbf{85.20} & \textbf{3.00}
& \textbf{99.50} & \textbf{98.70} & \underline{3.42}
& \textbf{61.07} & \textbf{91.28} & \textbf{75.89} \\
\midrule

\multicolumn{13}{c}{\textbf{Llama-3.1-8B-Instruct},
$B_{\mathrm{total}}=1024L$, context length $=128K$} \\
\midrule
Full KV
& 97.40 & 97.80 & 95.20 & 96.20 & 87.00 & 63.20
& 95.85 & 94.95 & 0.06 & 64.73 & 80.08 & 79.32 \\
SnapKV
& 93.60 & 93.80 & 1.40 & 85.80 & 25.60 & 0.00
& 76.80 & 27.00 & 0.08 & 26.49 & 71.76 & 45.67 \\
LAQ
& \underline{96.50} & \underline{97.00} & \underline{55.90}
& \underline{92.90} & \underline{63.40} & \underline{0.90}
& \underline{89.84} & \underline{73.31} & \underline{1.43}
& \underline{36.28} & \underline{75.20} & \underline{62.06} \\
SpecKV
& 95.20 & 95.30 & 29.90 & 89.40 & 45.40 & 0.30
& 83.55 & 50.93 & 0.86 & 31.51
& 73.52 & 54.17 \\
LookaheadKV
& 96.00 & 96.40 & 46.70 & 91.70 & 57.00 & 0.60
& 87.63 & 65.39 & 1.26 & 34.57
& 74.61 & 59.26 \\
\textbf{DeferKV}
& \textbf{97.40} & \textbf{98.80} & \textbf{78.40}
& \textbf{96.20} & \textbf{79.00} & \textbf{1.80}
& \textbf{95.25} & \textbf{93.10} & \textbf{1.68}
& \textbf{40.87} & \textbf{76.96} & \textbf{69.04} \\
\bottomrule
\end{tabular}%
}
\caption{Detailed RULER results across context lengths from 4K to 128K
using Llama-3.1-8B-Instruct with
$B_{\mathrm{total}}=1024L$.}
\label{tab:ruler-detail}
\end{table*}

\section{Results on Needle-in-a-Haystack}
\label{app:niah}

\begin{figure*}[t]
    \centering
    \includegraphics[width=\textwidth]{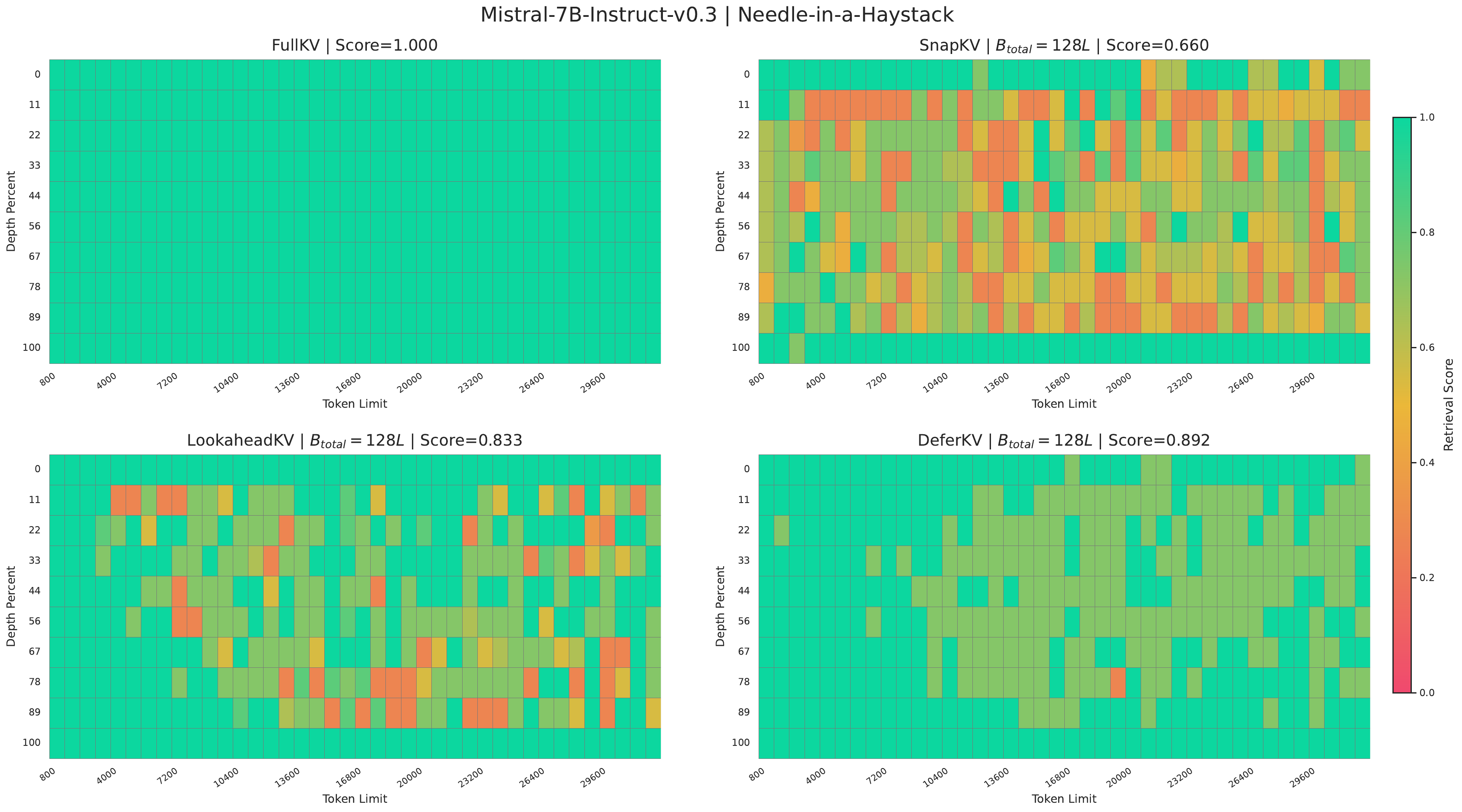}
    \caption{Needle-in-a-Haystack retrieval performance of Full KV, SnapKV,
    LookaheadKV, and DeferKV on Mistral-7B-Instruct-v0.3 with
    $B_{\mathrm{total}}=128L$.}
    \label{fig:niah-mistral-128}
\end{figure*}

\begin{figure*}[t]
    \centering
    \includegraphics[width=\textwidth]{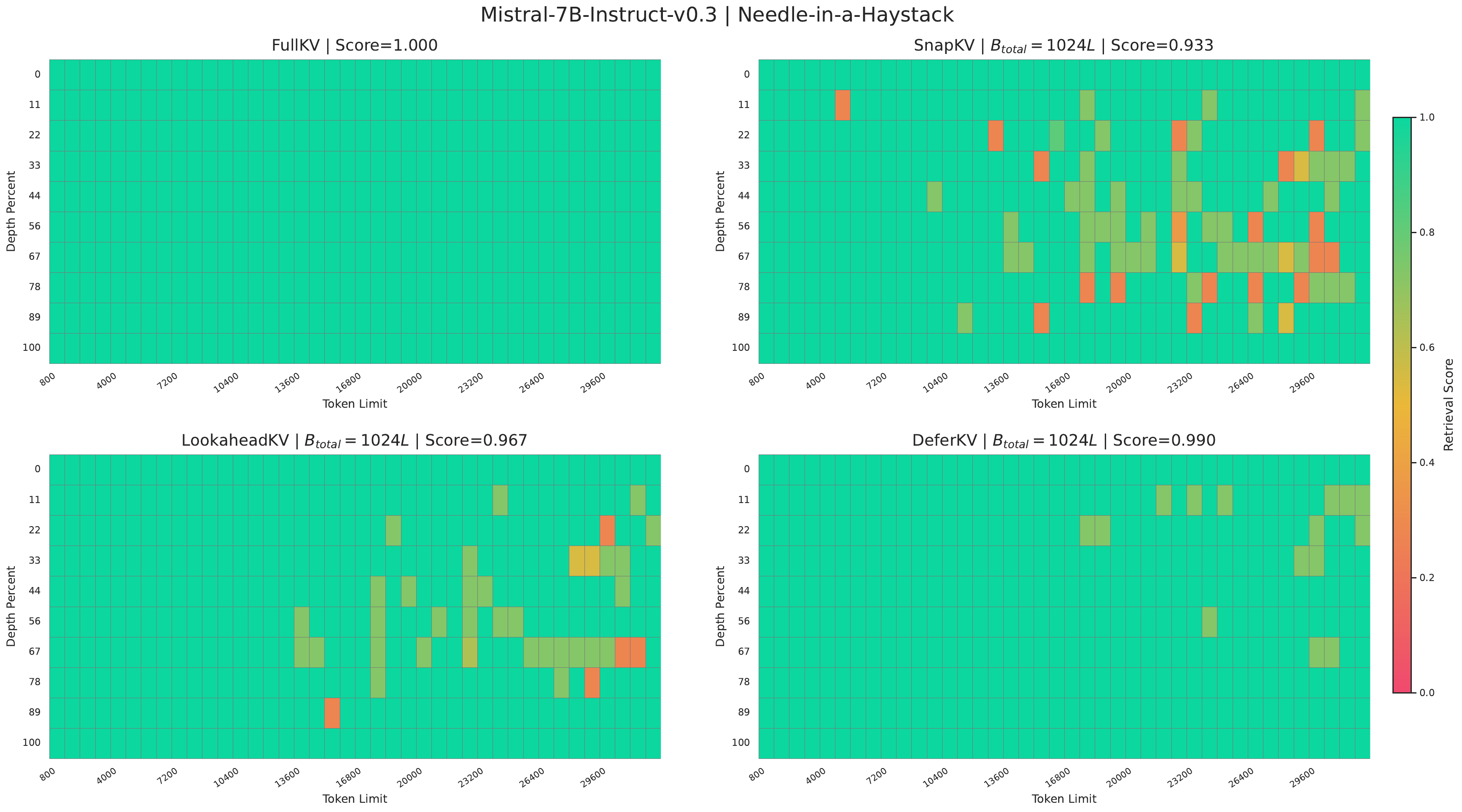}
    \caption{Needle-in-a-Haystack retrieval performance of Full KV, SnapKV,
    LookaheadKV, and DeferKV on Mistral-7B-Instruct-v0.3 with
    $B_{\mathrm{total}}=1024L$.}
    \label{fig:niah-mistral-1024}
\end{figure*}

In this section, we further evaluate the effectiveness and robustness of
DeferKV on the Needle-in-a-Haystack test using
Mistral-7B-Instruct-v0.3. This benchmark assesses a model's ability to
retrieve critical information embedded within long contexts. We compare
Full KV, SnapKV, LookaheadKV, and DeferKV under two cache budgets,
$B_{\mathrm{total}}=128L$ and $B_{\mathrm{total}}=1024L$.
Figures~\ref{fig:niah-mistral-128} and
\ref{fig:niah-mistral-1024} present the retrieval results under these two
settings, respectively. Under the constrained cache budget of
$B_{\mathrm{total}}=128L$, DeferKV achieves a mean retrieval score of
$0.892$ and outperforms LookaheadKV. These results provide a more
comprehensive comparison of DeferKV with existing methods across different
cache budgets and further demonstrate its effectiveness in preserving KV
pairs relevant to subsequent long-context retrieval.

\section{Numerical Analysis of Compression Delay Steps}
\label{app:delay-steps}

The results reported in the main paper show that increasing the compression delay
provides only modest performance improvements. Compared with $d=1$, using
$d=2$ increases the task score from 27.06 to 27.11, corresponding to a
relative improvement of only $0.18\%$, while end-to-end latency increases
from 11.562 to 11.618 seconds, or $0.48\%$. Therefore, waiting for a second
decode query introduces an additional full-cache decoding step but produces
only a negligible improvement in the final task score.

The same trade-off remains as the delay is increased further. At $d=4$, the
score improves by only $0.55\%$ relative to $d=1$, whereas latency increases
by $1.63\%$. Peak memory remains nearly unchanged across the evaluated
settings. These results suggest that additional decode observations can
slightly refine KV importance estimation, but the improvement is not large
enough to justify postponing compression. We therefore use $d=1$, which
incorporates real decode feedback at the earliest available step while
avoiding unnecessary full-cache decoding overhead.

\section{Effect of Stochastic Decoding}
\label{sec:stochastic-decoding}

DeferKV uses the first real decode query to estimate prompt-KV importance.
Since stochastic decoding may change the first generated token and its
corresponding query representation, we further examine whether DeferKV remains
effective under different decoding strategies. Specifically, we compare
greedy decoding with sampling temperatures $T\in\{0.2,0.8\}$ on LongBench
using Qwen3-8B with $B_{\mathrm{total}}=128L$. The same prompts, cache budget,
and method configurations are used across all decoding settings.

\begin{table*}[t]
\centering
\setlength{\tabcolsep}{7pt}
\renewcommand{\arraystretch}{1.05}
\begin{tabular}{@{}lcccccc@{}}
\toprule
\textbf{Decoding}
& \textbf{Full KV}
& \textbf{SnapKV}
& \textbf{LAQ}
& \textbf{SpecKV}
& \textbf{LookaheadKV}
& \textbf{DeferKV} \\
\midrule
Greedy
& 49.77
& 42.95
& 45.07
& 43.32
& 44.63
& \textbf{46.12} \\
$T=0.2$
& 49.64
& 42.07
& 44.58
& 40.94
& 43.02
& \textbf{46.01} \\
$T=0.8$
& 48.91
& 41.34
& 43.82
& 40.16
& 42.28
& \textbf{45.32} \\
\bottomrule
\end{tabular}
\caption{Average LongBench performance under greedy and stochastic decoding
using Qwen3-8B with $B_{\mathrm{total}}=128L$.}
\label{tab:stochastic-decoding}
\end{table*}

As shown in Table~\ref{tab:stochastic-decoding}, sampling with a low
temperature produces results close to those obtained with greedy decoding.
Increasing the temperature to $T=0.8$ leads to a moderate performance decrease
across all methods, reflecting the greater variability introduced during
generation. Nevertheless, DeferKV maintains strong performance relative to
the other KV compression methods under both temperatures. These results
indicate that its improvement is not limited to deterministic decoding and
that the decode-aware importance estimates remain effective when the first
generated token is sampled.

\end{document}